\documentclass[12pt]{article}
\RequirePackage{amsthm,amsmath}
\usepackage{natbib}
\usepackage{bm}
\usepackage{amsmath}
\usepackage{amssymb}
\usepackage{amsfonts}
\usepackage{multirow}
\usepackage{subfig}
\usepackage{graphicx}
\usepackage{xcolor}
\newcommand{\error}{\varepsilon} 
\newcommand{\bn}{g_n} 
\newcommand{\wn}{b_n}

\def\T{{ \mathrm{\scriptscriptstyle T} }}
\newcommand{\errorn}{\varepsilon_{n,  i}}
\newcommand{\traindat}{Z_{n_1}}

\newcommand{\testX}{\bm{X}_n^{(2)}}
\newcommand{\predn}{\bm{X}_{n,  i}}
\newcommand{\prednone}{\bm{X}_{n,  n_1 + 1}}
\newcommand{\resn}{Y_{n,  i} }
\newcommand{\bet}{\theta}

\usepackage{bbm}

\newtheorem{corollary}{Corollary}

\newcommand{\aln}{\alpha_{n_1}} 

\newtheorem{condition}{Condition}
\newtheorem{theorem}{Theorem}
\newtheorem{definition}{Definition}
\newtheorem{proposition}{Proposition}

\usepackage{mathtools}

\DeclarePairedDelimiter\floor{\lfloor}{\rfloor}

\graphicspath{{figures/}}
\begin{document}



  \title{\bf 
  Targeted Cross-Validation}
 \author{Jiawei Zhang\\
    School of Statistics, University of Minnesota\\
    zhan4362@umn.edu\\
    and \\
   Jie Ding \\
    School of Statistics, University of Minnesota\\
    dingj@umn.edu \\
    and\\
    Yuhong Yang\\
    School of Statistics, University of Minnesota\\
    yangx374@umn.edu}
\date{}
  \maketitle

\bigskip
\begin{abstract}
In many applications, we have access to the complete dataset but are only interested in the prediction of a particular region of predictor variables. A standard approach is to find the globally best modeling method from a set of candidate methods.
However, it is perhaps rare in reality that one candidate method is uniformly better than the others. 
A natural approach for this scenario is to apply a weighted $L_2$ loss in performance assessment to reflect the region-specific interest. We propose a targeted cross-validation (TCV) to select models or procedures based on a general weighted $L_2$ loss.  We show that the TCV is consistent in selecting the best performing candidate under the weighted $L_2$ loss. Experimental studies are used to demonstrate the use of TCV and its potential advantage over the global CV or the approach of using only local data for modeling a local region. 

Previous investigations on CV have relied on the condition that when the sample size is large enough, the ranking of two candidates stays the same. 
However, in many applications with the setup of changing data-generating processes or highly adaptive modeling methods, the relative performance of the methods is not static as the sample size varies.
Even with a fixed data-generating process, it is possible that the ranking of two methods switches infinitely many times.
In this work, we broaden the concept of the selection consistency by allowing the best candidate to switch as the sample size varies, and then establish the consistency of the TCV. This flexible framework can be applied to high-dimensional and complex machine learning scenarios where the relative performances of modeling procedures are dynamic.

\end{abstract}

\noindent%

\vfill

\section{Introduction}

Cross-validation (CV) is one of the most powerful tools for selecting models or procedures.   
Different CV methods include leave-one-out  \citep{allen1974relationship, stone1974cross,  geisser1975predictive}, leave-$p$-out 
\citep{shao1993linear, zhang1993model}, $V$-fold  \citep{geisser1975predictive}, repeated learning testing \citep{breiman2017classification, burman1989comparative, zhang1993model}, Monte-Carlo CV \citep{picard1984cross}, and generalized CV \citep{craven1978smoothing}.

There are various works related to the asymptotic properties of  CV.  For model selection,
the consistency of   CV  in both linear regression and time series models has been studied, e.g.,  \cite{li1987asymptotic}, \cite{shao1993linear,shao1997asymptotic}, and \cite{racine2000consistent}. For a broader setting of selecting general models or modeling procedures, \cite{yang2006comparing, yang2007consistency} provided conditions for
 consistency of CV in the context of regression and classification. The application of CV to selecting a model selection procedure in a high-dimensional setting was studied in \cite{zhang2015cross}. Apart from CV, the methods from \cite{baraud2011estimator} and \cite{baraud2014estimator} can also be used to select general modeling procedures for function estimation. It has been shown that CV can select tuning parameters for optimal nonparametric estimations such as the Nadaraya-Watson estimator  \citep{wong1983consistency}, smoothing spline \citep{craven1978smoothing, speckman1985spline}, and nearest neighbor  method \citep{li1984consistency}.  A comprehensive summary about CV-related works can be found in \citep{arlot2010survey,ding2018model}.  More recently, \cite{arlot2011segmentation} designed a CV-based method to detect change points in the heteroscedastic framework. 
A closed-form expression of risks from the leave-$p$-out CV, which provides insights into the choice of $p$ in both estimation and identification problems, was derived in  \cite{celisse2014optimal}. A non-asymptotic oracle inequality for the $V$-fold CV was obtained in \cite{arlot2016choice} , and it shows that the $V$-fold CV is asymptotically optimal when $V\rightarrow\infty$ in a nonparametric setting. A classification procedure that combines CV with aggregation was introduced in \cite{maillard2017cross}. A consistent CV procedure for selecting high-dimensional generalized linear models was studied in \cite{Feng2019}. A CV-based method that selects a subset of candidate models containing the best one with high probability was introduced in \cite{Lei2019}.

In various applications, we may want to select a candidate method with the best performance in a small region of interest. 
One may attempt to build a model for the region alone. However, this may not be a good solution because the local data are often limited
for efficient estimation, and the full data may help a good candidate model or procedure achieve optimal or near-optimal performance.
The current studies about  CV focus on selecting the candidate method with the best global performance and do not apply to finding the best one for a specific region. An effort was made in \cite{yang2008localized}, where the issue of selecting the best candidate at a single point has been studied, and the data-generating model is considered fixed.

In this work,  we propose a method named targeted CV (TCV) for the above problem of selecting an optimal candidate method for any particular region of interest. Our method allows for the consideration of flexible high-dimensional regression methods as candidates. More generally, it incorporates a weight function to reflect where the comparison of the candidate models or procedures is of most interest. A straightforward way for the weight is to assign a  0/1 value according to whether an observation is in the region. Compared with CV without weight, which selects the best global model, the TCV may work better than the regular CV when the globally best candidate does not perform uniformly the best. Additionally, the TCV can be used to compare the methods applied to the complete dataset with those based on the local data only.
On the one hand, candidate methods based on the complete dataset have the advantage of a larger sample size. On the other hand, the data outside the local region may introduce undesirable biases.
Fortunately, the TCV provides a data-adaptive way to compare them.

For the intended application of our TCV, the candidates are allowed to include non-model-based procedures in addition to models. We have found an apparently ignored but important aspect in previous theoretical developments on selection consistency when comparing general learning procedures. The issue is that the existing results assume that one procedure stays the best as the sample size approaches infinity. However, this view ignores the fact that in many applications, the comparison of the competing procedures is dynamic. Specifically, the performance ranking of the candidate procedures may keep changing as the sample size varies, especially when the candidate methods are highly adaptive and evolving with the sample size. 
We will provide an example to show that even with a fixed data-generating process, the ranking of two sensible models changes infinitely many times.
To accommodate for this inevitable complication in reality, we enable the TCV to work for a triangular array setup where the best candidate method may not be fixed. Under this broad setting, the goal is to find out the best candidate method under the current sample size. To define the best candidate in an asymptotic sense, we introduce a new concept of performance comparison elaborated in Section~\ref{neg}.

The outline of the paper is given below.
Section~\ref{set} defines the problem. Section~\ref{neg} introduces new concepts for performance comparison. Section~\ref{weightedcv} introduces our TCV and shows its consistency,  Section~\ref{simreal} presents the numerical studies. Section~\ref{conclusion} concludes the paper. The appendixes include the proofs. 

\section{Problem}\label{set}
We consider the random design regression model
$
Y_i = f(\bm{X}_i) + \error_i,
$
where $1\leq i \leq n$. Predictors $\bm{X}_{i} = (X_{i}^{(1)}, \cdots X_{i}^{(p)})$ are independent and identically distributed $p$-dimensional random variables.  For each 
$k\in \{1,2,\cdots, p \}$, $X_{i}^{(k)}$ can be either continuous  or  discrete. Let
$P_{\bm{X}}$ denote the joint probability distribution of the predictors and $\mathcal S$ denote the domain of $\bm{X}$. Let 
$\error_1,\cdots,\error_n$ be independent random errors with $E(\error_i|\bm{X}_i) = 0$ and $E(\error_i^2|\bm{X}_i)<\infty$ almost surely.
Let $W_n(\bm{x})$ be a nonnegative weight function that satisfies 
\begin{equation}\label{normeq}
\int_{\mathcal S} W_n(\bm{x})P_{\bm{X}}(d\bm{x})=1.
\end{equation}
 We define the weighted $L_{q}$ norm 
$
    \|f \|_{q,W_n}=
 \left  ( \int_{\mathcal S} W_n(\bm{x})\cdot |f(\bm{x})|^qP_{\bm{X}}(d\bm{x})\right)^{1/q},
$
where $0<q<\infty$. We require that 
$
  \| f \|_{2,W_n}<\infty \text{ and }
 \text{ess-sup}|W_n|<\infty.
$

We have a set of candidate regression  procedures 
$\delta_j\in\mathcal M$, where $j\in \mathcal J = \{1,2,\dots,m\}$, 
   and each candidate is based on a regression model or a general procedure. The set $\mathcal M$ may possibly change with $n$, but the number of candidate methods $m$ is upper bounded by a fixed constant. Let $\widehat{f}_{n}^{(j)}$ be the fitted regression function from $\delta_j$ with training data size $n$.
We want to find the candidate $\delta_{j^*}$ with $j^* = \underset{j \in \mathcal J}{\arg\min} \| f - \widehat{f}_{n}^{(j)}\|_{2,W_n}$.
Three examples of the weight function are as follows.  
\begin{description}
\item{Example 1. (Region-based weight)} Suppose we are only interested in the performances of the candidate methods  in a fixed region  $A$. We can take an indicator weight 
$W_n(\bm{x}) = C^{-1} \mathbbm{1}(\bm{x}\in A) $ where $C$ equals the probability of $\bm{X}\in A$ and $\mathbbm{1}(\cdot)$ is the indicator function.
\item{Example 2. (Conditional variance-based weight)} Suppose we know that the conditional variance of the response satisfies $Var(Y_i|\bm{X}_i)= Var(\epsilon_i)=\sigma^2(\bm{X}_i) $, where $ \sigma^2$ is a positive
function of predictors such that \\$\int_{\mathcal S} 1/\sigma^2(\bm{x})P_{\bm{X}}(d\bm{x})<\infty$. We can adjust the $L_2$ loss according to the conditional variance by
\begin{equation*}
W_n(\bm{x}) = \frac{ 1/\sigma^2(\bm{x}) }{\int_{\mathcal S} 1/\sigma^2(\bm{x})P_{\bm{X}}(d\bm{x})}.
\end{equation*}

\item{Example 3. (Single point-based weight)} Suppose we are interested in modeling $f(\bm{x})$ evaluated at a single point $\bm{x} = \bm{x}^*$. We can define the weight by a positive function centered at $\bm{x}^*$ that shrinks
toward that point as the sample size $n$ goes to infinity.
 For instance,
 \begin{equation*}
 W_n(\bm{x}) = \frac{ \exp{\left(-\| \bm{x} - \bm{x}^*\|^2_2\cdot n \right)}}{
\int_{\mathcal S} \exp{\left(- \| \bm{x} - \bm{x}^*\|^2_2\cdot n \right)P_{\bm{X}}(d\bm{x})}
 }.
 \end{equation*}
\end{description}

\section{A Neglected Aspect in Selection Consistency Theories for CV}\label{neg}
 In Section~\ref{Subsec_tria},  we address the need to extend the scope of the CV  to a more flexible framework that is based on the triangular array setting with possibly changing data-generating distributions. We also present the definition of the best candidate method under the extended setting. In Section~\ref{lnexp}, we exemplify the choice of splitting ratios needed to identify the best candidate  method.

\subsection{A triangular array setting}\label{Subsec_tria}
The selection consistency of CV has been established both for parametric model selection and for procedure selection. 
For the former, the candidates are assumed to be fixed linear models \citep{shao1993linear}. As the sample size increases to infinity, there is no ambiguity in terms of which model is the best. For the latter, results that allow the inclusion of general regression procedures are given in  \cite{yang2007consistency} and  \cite{zhang2015cross}. A major limitation of these theoretical results is that they assume a static ordering of the candidate procedures in terms of performance. In many applications, however, the relative performances among the candidate procedures may be dynamically changing with the sample size even if the true data-generating process stays fixed (see Section~\ref{Subsec_example}). Moreover, modeling methods for high-dimensional data usually assume changing true sparsity or true coefficients in deriving theoretical properties \citep{fan2004nonconcave, zhang2010nearly}, which implicitly indicate a triangular array setup. 

Now, consider two variable selection methods, and the goal is to choose between them. Suppose they are known to perform optimally under severe or moderate sparsity, respectively (e.g., the number of non-zero coefficients being of order $\log n$ and $n^{1/10}$, respectively). In this context, for an interesting theoretical investigation, it makes most sense to allow the unknown data generating model and the best candidate to change according to the sample size. This is just one example of situations where the relative ranking  is not static due to the fact that the modeling procedures may react quite differently with more or fewer observations. The present CV framework unfortunately cannot handle this reality.
Thus, it is essential to explicitly set up a flexible framework that gives each candidate method a chance to work better and confront the reality of possibly changing relative performances. Otherwise, a fixed truth or a rigid triangular array setup may lead to the conclusion that one of the two methods would always be preferred when $n$ is large enough, which is detached from many real applications. The above reasons motivate us to study the consistency of the TCV under the following triangular array framework, where data are represented in a triangular array. 


\begin{equation*}
\begin{split}
&(\bm{X}_{1, 1},Y_{1, 1})\\
&(\bm{X}_{2, 1},Y_{2, 1}), (\bm{X}_{2, 2},Y_{2, 2})\\
&\cdots \cdots \cdots \cdots \cdots \cdots \cdots \cdots \cdots \\
&(\bm{X}_{n, 1},Y_{n, 1}), (\bm{X}_{n, 2},Y_{n, 2}), \cdots, (\bm{X}_{n,  n},Y_{n,  n})\\
&\cdots \cdots \cdots \cdots \cdots \cdots \cdots \cdots \cdots \cdots \cdots \cdots\cdots \cdots .
\end{split}
\end{equation*}
In each row, the random pairs are i.i.d. For each $n$,  $\bm{X}_{n,  1},\cdots \bm{X}_{n,  n}$ follow the distribution $P_{\bm{X}_n}$, and the weighted
$L_2$ norm is defined according to $P_{\bm{X}_n}$. Under this setting, the goal is to find out the best candidate method given the current sample size.

The fitted regression function $\widehat{f}_{n_1}^{(j)}$ with $j\in\mathcal J$ was obtained by first sampling (without replacement) $n_1$ observations as the training set and then applying the candidate methods $\delta_j$ with $j\in\mathcal J$ to this dataset.
Let  $ c_{(n_1,n)}$ be  a sequence of positive numbers. Let $l_n$ be a sequence of positive  integers such that $l_n < n$, and  $l_n \rightarrow \infty$ as $n\rightarrow \infty$. For illustrative purposes, we first focus on the case with two candidate methods. Let $\bn$ and $\wn$ be two sequences that  take values from $\mathcal J = \{1,2\}$ and satisfy
$\bn + \wn = 3$, where $\bn$ stands for ``good'' and $\wn$ for ``bad''.

\begin{definition}[$(W_n, l_n, c_{(n_1,n)})$-better]\label{def_better} The candidate method $\delta_{\bn}$ is said to be the $(W_n, l_n, c_{(n_1,n)})$-better one asymptotically out of the two candidates $\delta_1$, $\delta_2$  at sample size $n$  if for all $l_n \leq n_1 <n$, we have
\begin{equation}\label{eq:bett}
P(\| f-\widehat{f}_{n_1}^{(\wn)} \|_{2,W_n}\geq(1+c_{(n_1,n)})\| f- \widehat{f}_{n_1}^{(\bn)}\|_{2,W_n})\rightarrow 1,
\end{equation}
as $n\rightarrow\infty$.
\end{definition}

The quantities involved in the above definition capture the key aspects in the TCV comparison. The sequence $c_{(n_1,n)}$ characterizes the difference between the losses of the two candidate methods. It is related to $n_1$ in the sense that the convergence rate of the fitted regression function depends on the training data size. It may also depend on $n$ since the data-generating process, weight $W_n$, and better candidate $\bn$ may change with $n$. In the high-dimensional setting with the number of predictor variables increasing to infinity, $c_{(n_1,n)}$ may need to go to $0$ when the two candidate methods are very close.  An example concerns the choice of $c_{(n_1,n)}$ for comparing the underlying true model with an over-fitting model with one additional term.
As will be seen in our main theorem, the TCV will require a higher portion of the test data to handle the challenge of a decreasing performance difference in such a case. If $l_n$ is much smaller than $n$, then for a wide range of choices of $n_1$ for the TCV, the comparison 
result of $\delta_{\bn}$ and $\delta_{\wn}$ at the reduced sample size $n_1$ matches that 
at the full sample size. In contrast, suppose for instance, $\delta_{\bn}$ is $(W_n, l_n, c_{(n_1,n)})$-better than $\delta_{\wn}$, but $\delta_{\bn}$ is not better even at a slightly reduced sample size
$n_1 < n$, then one may not be able to tell which candidate is better at the full sample size $n$ when data splitting is done. In this case,  it is hard to get a consistent selection for the TCV, as expected.

In the general case where $\mathcal M$ may contain more than two candidate methods, we let $\bn$ denote the to-be-defined best candidate and 
\begin{equation}\label{eq_def_J}
\mathcal J_b\triangleq \{j\in \mathcal J : j\neq \bn\}.	
\end{equation} 
We define the following extension of Definition~\ref{def_better}. 
\begin{definition}[$(W_n, l_n,  c_{(n_1,n)})$-best]\label{def_bestdef} A candidate method $\delta_{\bn}$ from $\mathcal M$ is said to be the $(W_n, l_n,$ $c_{(n_1,n)})$-best one asymptotically if there exist $0<l_n<n$ and $c_{(n_1,n)}>0$, such that the method
$\delta_{\bn}$ is $(W_n, l_n, 
c_{(n_1,n)})$-better than $\delta_{i}$  for each $i\in \mathcal J_b$.
\end{definition}

\subsection{An example of the changing $l_n$}\label{lnexp}
Recall that under the $(W_n, l_n, c_{(n_1,n)})$-better condition, $\delta_{\bn}$ is better than $\delta_{\wn}$ as long as the training set size $n_1$ is larger than or equal to a lower bound $l_n$. In this subsection, we provide a toy example of $l_n$, which shows that  the increasing speed of $n_1$ cannot be too slow compared with $n$, in order to guarantee that the performance ranking of the candidate methods at the sample size $n_1$ remains the same as that of $n$.

We consider the data-generating process 
$
Y_i = f(X_i) + \varepsilon_i,
$
where $f(x) = x^{2}$, $\varepsilon_i$'s are i.i.d.\ from  $N(0, \sigma^2)$,
and  $X_i$'s are i.i.d.\ from $U(0,1)$.
We are interested in estimating $f(x)$ where $x$ is close to 0.
We consider
\begin{equation*}
W_n = \begin{cases}
C^{-1} & \text{if}\  0\leq x \leq  n^{-\frac{1}{8}},\\
0& \text{otherwise},
\end{cases}
\end{equation*}
where the normalizing constant $C = \int^{n^{-{1}/{8}}}_0 dx = n^{-{1}/{8}}$.
The candidate models are
\begin{align*}
\text{Model 1: } \widehat{f}_{n_1}^{(1)}(x) =&\  x^{2} + \frac{1}{n_1}\sum^{n_1}_{i=1}(y_i - x_i^2) =f(x) + \frac{1}{n_1}\sum^{n_1}_{i=1}\varepsilon_i,\\
\text{Model 2: } \widehat{f}_{n_1}^{(2)}(x)  \equiv&\  0.
\end{align*}
For model 1, we have
\begin{equation*}
\| f- \widehat{f}_{n_1}^{(1)}\|_{2,W_n}^2 
=\int^{n^{-\frac{1}{8}}}_0 \left(f(x) - f(x) -\frac{1}{n_1} \sum^{n_1}_{i=1}\varepsilon_i\right)^2 \cdot C^{{-1}} dx
=\left(\frac{1}{n_1}\sum^{n_1}_{i=1}\varepsilon_i\right)^2.
\end{equation*}
Therefore,
$\| f- \widehat{f}_{n_1}^{(1)}\|_{2,W_n}^2 $
converges at the rate $n_1^{-1}$. For model 2, 
\begin{equation*}
\| f- \widehat{f}_{n_1}^{(2)}\|_{2,W_n}^2 
=\int^{n^{-\frac{1}{8}}}_0 (f(x) -0)^2\cdot C^{{-1}}dx
=\int^{n^{-\frac{1}{8}}}_0 x^4dx \cdot n^{\frac{1}{8}}
= \frac{1}{5}\cdot n^{-\frac{1}{2}}.
\end{equation*}
Thus, $\| f- \widehat{f}_{n_1}^{(2)}\|_{2,W_n}^2 $ converges 
exactly at rate $n^{-\frac{1}{2}}$, regardless of $n_1$.

For the estimators $\widehat{f}_{n}^{(1)}$ and $\widehat{f}_{n}^{(2)}$ using the complete dataset, the convergence rates of $\| f- \widehat{f}_{n}^{(1)}\|_{2,W_n}^2$ and $\| f- \widehat{f}_{n}^{(2)}\|_{2,W_n}^2$ are $n^{-1}$ and $n^{-\frac{1}{2}}$, respectively.
Therefore, the weighted $L_2$ loss of  $\widehat{f}_{n}^{(1)}$ converges faster than the 
weighted $L_2$ loss of  $\widehat{f}_{n}^{(2)}$.
However, in order to get the same ranking of $\widehat{f}_{n_1}^{(1)}$ and $\widehat{f}_{n_1}^{(2)}$ based on the training data with size $n_1$, we need $n_1/\sqrt{n}\rightarrow\infty$ as $n\rightarrow\infty$.

The above example is meant for a quick illustration. In reality, suppose that $\delta_1$ and $\delta_2$ denote two teams of data scientists participating in an online data competition. It is conceivable that the two teams may try various learning tools with validation feedback on their performances, and their relative ranking may not be static. In this case, to fairly evaluate their performances at a reduced sample size $n_1<n$, the lower bound $l_n$ needs to be carefully chosen.

\section{Method and Main Result}\label{weightedcv}

In Section~\ref{Subsec_TCVDef}, we present the TCV method. In Section~\ref{main}, we introduce the theoretical result that shows the model selection consistency of the TCV. In Section~\ref{Subsec_example}, we present a nonparametric regression example with alternating best candidate method and verify the requirements for the property of the TCV. In Section~\ref{Subsec_mTCV}, we extend the theoretical result of the TCV from a single splitting to multiple splittings that aim to stabilize the selection result.

\subsection{Targeted cross-validation}\label{Subsec_TCVDef}

Recall that we have randomly partitioned the dataset into a training set with size $n_1$ and a test set with size $n_2 = n-n_1$. 
We define our weighted squared prediction error by
\begin{equation}\label{tcvdef}
TCV_{W_n}(\widehat{f}_{n_1}^{(j)})=\sum^n_{i=n_1+1}(Y_i-\widehat{f}_{n_1}^{(j)}(\bm{X}_i))^2\cdot W_n(\bm{X}_i),
\end{equation}
 where $j\in\mathcal J$ and the summation is taken over the test set, and for notational simplicity, $(\bm{X_i}, Y_i)$ denote $(\bm{X_{n,i}}, Y_{n,i})$.
The TCV selects the candidate $\hat{j} = \underset{j \in \mathcal J}{\arg\min}\ TCV_{W_n}(\widehat{f}_{n_1}^{(j)})$.  

\subsection{The main theorem}\label{main}
We first introduce some necessary definitions and conditions.
\begin{definition}[$W_n$-consistent selection] We assume that there exists a candidate method $\delta_{\bn}$ that is the $(W_n, l_n,  c_{(n_1,n)})$-best out of $\mathcal M$ at sample size $n$. A selection rule is called 
$W_n$-consistent if its probability of selecting $\delta_{\bn}$ goes to 1 as $n\rightarrow \infty$.
\end{definition}
\begin{definition}[Lower bound of the rate of the weighted $L_2$ loss]\label{def_lbL2} A fitted regression function $\widehat f_{n_1}$ is said to have $(W_n, l_n)$-convergence rate lower  bounded by $a_{(n_1,n)}$ under the weighted $L_2$ loss  if for each $ 0<\epsilon<1$, there exist $ {c}_\epsilon>0, N\in \mathbb Z^+$ such that for all $n\geq N$
and $l_n \leq n_1 \leq n$,
\begin{equation}
P\left(\| f-\widehat{f}_{n_1} \|_{2,W_n}\geq {c}_\epsilon a_{(n_1,n)}\right)\geq1-\epsilon.
\end{equation}
\end{definition}

%


\begin{condition}[Error variances]\label{cond_error_variance} The error variances $E(\error^2_i|\bm{X}_i)$ are upper bounded by a constant $\overline\sigma^2>0$ almost surely for all $i\geq 1$.
\end{condition}

\begin{condition}[Relating weighted $L_4$ and $L_2$ losses]\label{cond4} There exists a sequence of positive numbers $M_{n}$ such that for all $l_n\leq n_1 < n$,
$$\sup_{j\in\mathcal J}\bigl(\| f-\widehat{f}_{n_1}^{(j)}\|_{4,W_n}\big/\| f-\widehat{f}_{n_1}^{(j)}\|_{2,W_n}\bigr)=O_p(M_{n}),$$
as $n\rightarrow\infty$. 
\end{condition}
\begin{condition}[Lower bound of the convergence rates]\label{cond_lb_rate}
There exists a sequence $q_{(n_1,n)}$ such that
	for each $j\in\mathcal J_b$, $\widehat f_{n_1}^{(j)}$ has $(W_n, l_n)$-converge rate lower  bounded by $q_{(n_1,n)}$ under the weighted $L_2$ loss.

\end{condition}
Condition~\ref{cond_error_variance} is a mild requirement that is satisfied when, e.g., the random errors have the same finite variance.  For Condition~\ref{cond4}, it has been shown that for some familiar function classes, the estimators may have the same rates for both $L_4$ and $L_2$ losses. For instance, Lipschitz class~\citep[Section~1.2,][]{nemirovski2000topics}, H\"older class (see, e.g., Section~1.3 of \cite{nemirovski2000topics} and \cite{stone1982optimal}).  and Sobolev class~\citep[Section~2.1\&2.2,][]{nemirovski2000topics}. Also, an example that compares AIC-based and BIC-based selection procedures with $M_n \equiv 1$ can be found in Section~3 of \cite{zhang2015cross}. For Condition~\ref{cond_lb_rate}, similar to $c_{(n_1,n)}$,  both $n_1$ and $n$ are involved in determining the rate $q_{(n_1,n)}$. For instance, in the example from Section~\ref{lnexp}, when $n_1/\sqrt{n}\rightarrow\infty$ as $n\rightarrow\infty$, we have that $q_{(n_1,n)} = {1}/{\sqrt{n}}$.

Let $S_{W_n}$ denote $\underset{\bm{x}\in\mathcal S}{\text{ess-sup}}(W_n(\bm{x}))$. 
Our main theorem is as follows.
\begin{theorem}[$W_n$-consistency of the TCV]\label{thm} Assume that Conditions~\ref{cond_error_variance}-\ref{cond_lb_rate} hold, and the data splitting is such that for all $l_n\leq n_1< n$, we have
\begin{description}
\item{\textbf{\textit{(i).}}} $n_2\cdot c_{(n_1,n)}^2/(S_{W_n}M_{n}^{4})\rightarrow\infty$,
\item{\textbf{\textit{(ii).}}} 
$n_2\cdot (c_{(n_1,n)}\cdot q_{(n_1,n)})^2/S_{W_n}\rightarrow\infty$,\
\end{description}
as $n\rightarrow \infty$.
Then, the  TCV is $W_n$-consistent. 
\end{theorem}
The detailed proof can be found in Appendix~\ref{appendix_A}. The requirements \textbf{(\textit{i})} and \textbf{(\textit{ii})} indicate that the TCV needs the test size $n_2$ to be sufficiently large. In the case of
comparing nested high-dimensional regression models with a fixed number of additional predictors, under some mild conditions from Section~4.2 of \cite{zhang2015cross}, we have $(c_{(n_1,n)}\cdot q_{(n_1,n)})^2 = O_p(1/n_1)$. Then, if $M_{n} = 1$ and  both $S_{W_n}$ and $q_{(n_1,n)})$ are bounded above by  fixed constants, the requirements (\textbf{\textit{i}}) and (\textbf{\textit{ii}}) can be simplified by $n_2\rightarrow\infty$ and $n_2/n_1\rightarrow\infty$.
This requirement on the splitting ratio is in accordance with those from Theorem~1 of \cite{shao1993linear} for classical linear regression and Theorem~3.3 from \cite{celisse2014optimal} for density estimation. It is interesting to note that this splitting ratio direction may be opposite to that for risk prediction or asymptotically optimal estimation in some contexts, which need $n_2/n_1\rightarrow0$ (see, e.g., \cite{burman1989comparative, burman1990estimation, arlot2016choice}).
When $W_n(x)$ is chosen to focus on a region $R_n$ with decreasing probability, which implies that $S_{W_n}$ goes to infinity, the two conditions (\textbf{\textit{i}}) and (\textbf{\textit{ii}}) require $n_2$ to be larger compared with the global CV. In such cases, it is crucial to have enough evaluation data points in order to separate close competitors on a small region.

Compared with former related works, e.g., \cite{yang2007consistency} and \cite{zhang2015cross}, our theoretical results differ in three major aspects. First, our theory takes the effect of the weight into account. It applies to a broader range of CV applications such as illustrated by Examples~1 to 3 in Section~\ref{set}, and our result highlights the need to adjust the data splitting ratio accordingly. Second, our method works in a more general and realistic framework with possibly changing relative performances of the candidate methods. Third, with an improved derivation, we have removed the requirements on the upper bound of the sup-norm loss and exact rate of the $L_2$ loss as those in Condition~1 and Definition~3 of \cite{yang2007consistency} and Conditions~1 and 4 of \cite{zhang2015cross}.

\subsection{An illustrative example with an alternating better candidate method}\label{Subsec_example}
In this subsection, we present an example where we observe the alternating relative performances between two candidate methods. This example is particularly interesting in that the data-generating process is fixed. We also verify the conditions for the TCV and provide a valid range for the required data splitting ratio.

\begin{figure}[h]
\centering
\includegraphics[width=0.9\textwidth]{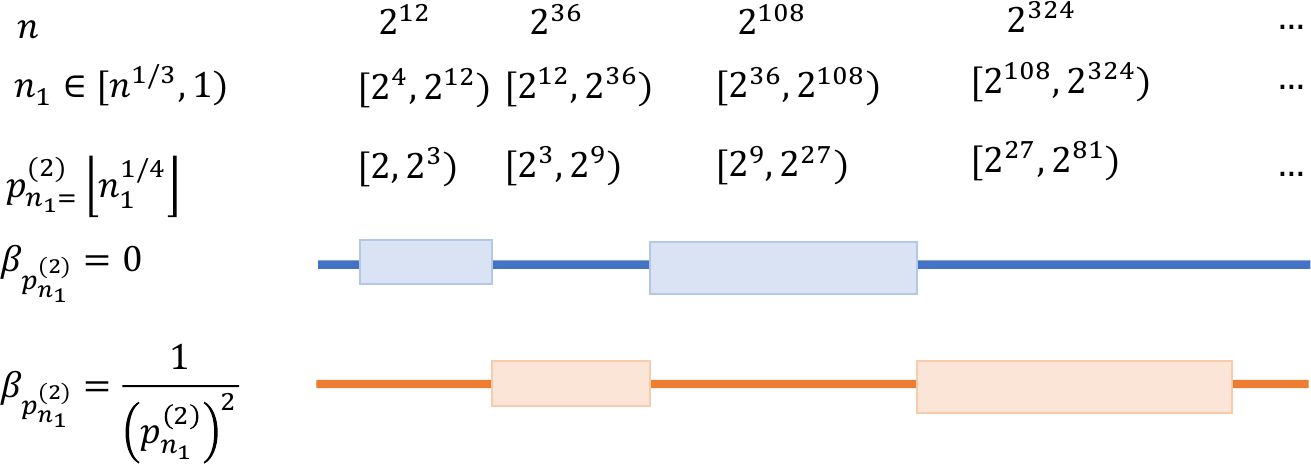}
\caption{An illustration of $\beta_{p^{(2)}_{n_1}}$ when $n_1\in[n^{1/3},n)$ and $n \in \mathbb S$.}
\label{Fig_betaInterval}
\end{figure}

We consider the i.i.d.\ data-generating process
$
y = f(x) + \epsilon,
$
where $f(x) = \sum_{j=1}^\infty \beta_j\phi_{j}(x)$, $x\sim U(0,1)$, the random error $\epsilon$, assumed to be independent of $x$, has mean zero and variance one, and for convenience,
$
\phi_{j}(x)= \sqrt{2}\sin 4^j\pi x.
$
Let $\mathbb S=\{2^{(12\times 3^{q-1})}:q\in \mathbb N \}$, where $\mathbb N$ denotes the set of natural numbers. Let $\mathbb S_1$ denote the subset of $\mathbb S$ with odd $q$ and $\mathbb S_2$ denote those with even $q$.
The data-generating coefficients are 
$$
\beta_j=
\begin{cases} 
0 &\text{when }j\in \cup_{\nu\in\mathbb S_1}[\floor{\nu^{1/12}},\floor{\nu^{1/4}}],\\
\frac{1}{j^2}	 &\text{otherwise},
\end{cases}
$$
where $\floor{x}$ stands for the largest integer no bigger than $x$.
Our candidate models with training set size $n_1$ are
\begin{align*}
\text{Model 1:}\quad\hat f^{(1)}_{n_1} =& \sum_{j=1}^{p^{(1)}_{n_1}}\hat\beta_j^{(n_1)}	\phi_{j}(x),\quad
\text{Model 2:}\quad \hat f^{(2)}_{n_1} = \sum_{j=1}^{p^{(2)}_{n_1}}\hat\beta_j^{(n_1)}	\phi_{j}(x),
\end{align*}
where $\hat\beta_j^{(n_1)} = \frac{1}{n_1}\sum_{i=1}^{n_1} y_i\phi_j(x_i)$, $p^{(1)}_{n_1} = \floor{n_1^{1/4}}-1$, and $p^{(2)}_{n_1} = \floor{n_1^{1/4}}$. This definition is valid if we consider $n_1\geq 16$.
The coefficient $\beta_{p^{(2)}_{n_1}}$, which corresponds to the additional variable in model 2 not in model 1, has an alternating pattern as shown in Figure~\ref{Fig_betaInterval}. We want to find out the better candidate model when $n \in\mathbb S$. 

\begin{proposition}[Alternating relative performance]\label{prop_1} Take $l_n= n^{1/3}$, $n_1\in [l_n,n)$, and $W_n(\bm{x})\equiv 1$. Then, 
when $n\in \mathbb S_1$, model 1 is 
$(W_n, l_n, n_1^{-1/3}/3)$-better than model 2, and when $n\in \mathbb S_2$, model 2 is $(W_n, l_n, n_1^{-1/3}/3)$-better than model 1.
\end{proposition}
 \begin{proposition}[Verifying the requirements for the TCV]\label{prop_2}	
Assume that $n_2/n_1^{17/12}\rightarrow\infty$ as $n\rightarrow\infty$ 
With $c_{(n_1,n)} = n_1^{-1/3}/3$ and $W_n(\bm{x})\equiv 1$, we have
\begin{description}
\item{\textbf{\textit{i.}}} $n_2\cdot c_{(n_1,n)}^2/M_{n}^{4}\rightarrow\infty$,
\item{\textbf{\textit{ii.}}} $n_2\cdot (c_{(n_1,n)}\cdot q_{n_1,n})^2\rightarrow\infty,$
\end{description}
as $n\rightarrow\infty$. That is, the requirements for consistency of the TCV in this example  are satisfied.
\end{proposition}
The proofs are in  Appendices~\ref{Appendix_prop1Proof} and \ref{Appendix_prop2Proof}.

\subsection{Multiple data splittings}\label{Subsec_mTCV}

In practice, we apply the TCV with multiple data splittings in order to lower the variability of the selection result. The need of a number of data splittings to achieve stability in the outcome of selection is numerically demonstrated in \cite{zhan2021profile}.
The training set and test set in each single splitting are assumed to be independent. With any multiple-splitting method, we obtain a set of MSEs for each candidate. Next, we introduce two ways to combine the results. Let 
$\widehat{f}_{n_1,k}^{(j)}$ with $j\in\mathcal J$ be the estimators from the candidate procedures based on the training set in the $k$-th splitting. Let $K$ denote the total number of splittings. We define
\begin{align}
	MTCV_{W_n}^a(\delta_j)&=\frac{1}{K}\sum^{K}_{k=1}TCV_{W_n,k}(\widehat{f}_{n_1,k}^{(j)}),\label{multisplit}\\
		MTCV_{W_n}^v(\delta_j)&=\frac{1}{K}\sum^{K}_{k=1}\mathbbm{1}(j = \arg\min_{i\in\mathcal J}TCV_{W_n,k}( \widehat{f}_{n_1,k}^{(i)})),\label{multisplit_v}
\end{align}
for the multiple splitting TCV by averaging and by voting, respectively. The two multiple splitting TCV methods will select  $\delta_j\in\mathcal M$ that minimizes $MTCV_{W_n}^a(\delta_j)$ and maximizes $MTCV_{W_n}^v(\delta_j)$, respectively.
The following corollary shows that the TCV with multiple splittings is also  $W_n$-consistent.

\begin{corollary}[Consistency of the TCV with multiple splitting]\label{cor_MTCV_consistency}
	Under the same conditions as in Theorem~\ref{thm}, if the number of splittings $K$ is independent of the data, the multiple splitting TCV by voting is $W_n$-consistent. If additionally, $K$ is upper bounded by a fixed constant (independent of $n$), the multiple splitting TCV by averaging is also $W_n$-consistent.
\end{corollary}

\section{Simulations and Real Data Example}\label{simreal}

The goal for the simulation and real data examples is to investigate if the TCV can indeed improve over  (regular) CV when one's focus is not on the global performance. We also include local-dataset-based methods as candidates in some cases. The examples are chosen to highlight the differences among the competing methods. For multiple splittings, we apply the scheme of Monte-Carlo CV and aggregate the results by averaging as shown in Equation \eqref{multisplit}.

\subsection{Simulation 1: a simple model versus a comprehensive model}\label{s1}
Consider an example that involves  the comparison between a simple model and a comprehensive model.

 Let the data-generating process  with predictors $X^{(0)},\dots, X^{(100)}$ and $\mathcal I$ be 
\begin{equation*}
\begin{split}
Y&=X^{(0)}+(1-\mathcal{I})(X^{(1)}+\cdots+X^{(100)})+\varepsilon,
\end{split}
\end{equation*}
where  $X^{(0)},\dots, X^{(100)}$ follow a multivariate normal distribution with mean $\bm{0}$ and covariance matrix
\begin{equation*}
 \left(
\begin{matrix}
20 & 0 & 0 &\ldots & 0\\
0& 0.1 &  0.1&\ldots & 0.1\\
0&0.1  &  0.1 & \ldots &0.1\\
\vdots &\vdots &  \vdots & \ddots & \vdots\\
0 &0.1  &   0.1    &\ldots & 0.1
\end{matrix}
\right),
\end{equation*}
$\mathcal I$ follows $Bernoulli(0.1)$ and is independent of $X^{(0)},\dots, X^{(100)}$, and
the random error $\varepsilon$ follows $N(0,\sigma^2)$. We consider error  variances $\sigma=25$ or $3$. 
The goal is to find out the best candidate method  for the local region 
\begin{equation}\label{loreg}
\{(x^{(0)}, x^{(1)}, x^{(2)}\cdots x^{(100)},  \mathcal I)^{\T}:\ \mathcal I = 1, (x^{(0)}, x^{(1)}, x^{(2)}\cdots x^{(100)})^\T\in \mathbb R^{101}\}.
\end{equation}
Our candidate models are: 
\begin{equation*}
\begin{split}
\text{$\delta_1$: }& Y = \beta_0 X^{(0)} + \varepsilon,\text{  fitted on the complete dataset},\\
\text{$\delta_2$: }& Y = \beta_0 X^{(0)}+\cdots + \beta_{100}X^{(100)} + \beta_{101}\mathcal I \cdot X^{(1)}
+\cdots \beta_{200}\mathcal I \cdot X^{(100)} + \varepsilon,\\
&\text{  fitted on the complete dataset},\\
\text{$\delta_3$: }& Y = \beta_0 X^{(0)} + \varepsilon,\text{  fitted on the local dataset, where $\mathcal I = 1$}.
\end{split}
\end{equation*}
The candidate models $\delta_1$ and $\delta_3$ are only correct in the local region in \eqref{loreg} and $\delta_2$ is the overall correct model. 
 Intuitively, $\delta_1$ can be better than $\delta_2$ in the local region since 
$\delta_2$ has too many parameters to be estimated.
As for $\delta_1$ versus $\delta_3$, it depends on whether the gain from the increased sample size outweighs the loss from the variability introduced by $X^{(1)},\dots,X^{(100)}$. Thus,  the two error variances $\sigma_1^2$ and $\sigma_2^2$ are likely to result in different outcomes. 

First, we randomly generate a dataset with $n=800$. Next, we apply both the CV  and TCV with $W_{n}(\bm{x})=\mathbbm{1}(\mathcal I = 1) $. The training set and test set sizes $n_1 = n_2 = 400$,  and the number of splittings $K = 100$. To measure the performance of the candidate models and selection results, we independently generate an evaluation set from the data-generating model with size 5000 and calculate \textbf{i)} weighted MSE with $0/1$ weight on $\mathcal I$, which measures the local performance, and \textbf{ii)} MSE without weight, which measures the overall performance,  for the three candidate models and the models selected by the CV and TCV.  The MSEs and associated standard errors based on $500$ replications are shown in Table~\ref{sim1_1}.

\begin{table}[!ht]
\centering
\caption{The MSEs from $\delta_1$, $\delta_2$, $\delta_3$, TCV, and CV for both the local and overall performances. The standard errors of the MSEs are shown in parentheses. The bold numbers stand for the better one out of the CV and TCV.}
\label{sim1_1}
\centering
\begin{tabular}{lllllll}\hline \noalign{\smallskip}
&&$\delta_1$ & $\delta_2$ & $\delta_3$ & TCV    & CV             \\ \noalign{\smallskip}\hline \noalign{\smallskip}
$\sigma = 25$&Local   & 62.8   & 65.6       & 63.3   & $\bm{63.1}$   & 65.6  \\
   &    & (0.2) & (0.2)      & (0.2) & (0.2) & (0.2) \\
&Overall  & 1528.3   & 631.7         & 1533.1   & 1518.0   & $\bm{631.7}$    \\
    &   & (1.4) & (0.6)      & (1.5) & (4.9) & (0.6)      \\\noalign{\smallskip}
$\sigma = 3$&Local & 0.997   & 0.945       & 0.911    & $\bm{0.914}$    & 0.945    \\
  &     & (0.007) & (0.004)    & (0.003) & (0.003) & (0.003) \\
&Overall   & 910.8    & 9.1          & 909.9      & 886.6      & $\bm{9.1}$        \\
&       & (0.861) & (0.009)     & (0.864)  & (6.474)  & (0.009) \\ \noalign{\smallskip}\hline
\end{tabular}
\end{table}

It can be seen from Table~\ref{sim1_1} that $\delta_2$ is not the best model for the local region. Therefore, in this case, the complex global model, while being correct, does not perform as well as the simple  model or that based  on the local region due to the relatively small sample size given the complexity of the whole model.
The table also shows that for $\sigma = 25$,  $\delta_1$ outperforms $\delta_3$ for the local region. This is because with the use of more observations, $\delta_1$ gains much in variance reduction and has a better performance than $\delta_3$ for the local region.  For $\sigma = 3$, however, the gain from the variance reduction is blown away by the bias introduced by the outside data (recall that $X^{(1)},\dots,X^{(100)}$ are not included in $\delta_1$), resulting in the worse local performance. In both cases, $\delta_2$ always has the best overall performance. 
For the selection results, the TCV has better performance than CV for the local region as desired, but worse for the overall performance.

To demonstrate the effect of the number of data splittings in TCV, we calculate the weighted cross validation MSEs (as defined in (\ref{multisplit})) with $K=1$ and $K=100$ respectively. As shown in the left panel of Table~\ref{sim1_splits}, 
for the weighted cross validation MSEs of the candidate models,  the respective standard deviations based on 500 replications decrease with the increased number of splittings.  The same is true for the weighted squared $L_2$ loss (simulated based on 5000 independently generated predictor values) of the regression estimators from the models selected by TCV, as seen in the right panel of Table~\ref{sim1_splits}. 
\begin{table}[htbp]
  \small
  \centering
  \caption{Effect of the number of data splittings ($K=1$ versus $100$) on cross validation errors and regression estimation losses.}\label{sim1_splits}
  \subfloat[Standard deviations of the weighted cross validation MSEs of the candidate models.]{
  \begin{tabular}{llll}\hline
Candidate Model & $\sigma$ & $K=1$ & $K=100$\\ \hline
$\delta_1$ & 25 & 15.07   & 10.61   \\
       & 3  & 0.37  & 0.21    \\
$\delta_2$ & 25 & 8586.74  & 5724.64 \\
       & 3  & 123.65   & 82.43   \\
$\delta_3$ & 25 & 7700.46  & 5368.10 \\
       & 3  & 110.89   & 77.30  \\\hline
\end{tabular}      }\hspace{1cm}
  \subfloat[Standard deviations of the weighted squared $L_2$ losses of the models selected by TCV.]{%
  
  \begin{tabular}{lllll}\hline
   &   $\sigma$      & $K=1$ & $K=100$      \\\hline
Local&25  &5.25	 &	4.93\\
  &3 & 0.089		&0.073\\
Overall &25 & 323.25 &86.60\\
  &3 & 372.85 &	154.93 \\\hline\end{tabular}
  }
\end{table}

\subsection{Simulation 2: a continuous weighting for TCV}\label{simulation2}
Sometimes, it may be preferable to assign weights between $0$ and $1$, e.g., $W(\bm{x})=a$ if $\bm{x}$ is in a certain region, where $0<a<1$ and $W(\bm{x})=1 - a$ otherwise. In this way, we aim to find out the candidate methods with excellent performance for the region with the larger weight value and acceptable performance on the remaining part. 

\begin{figure}[!ht] \centering\includegraphics[width=0.8\textwidth]{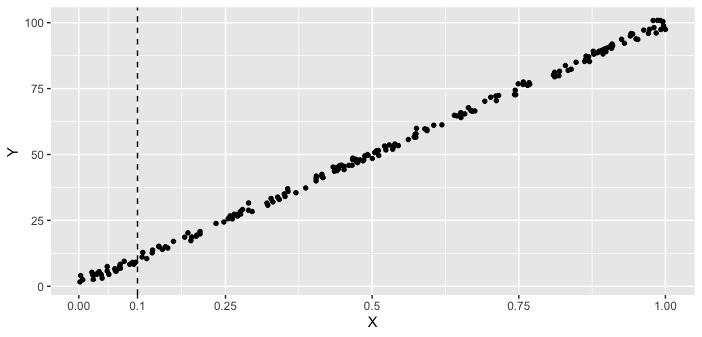}\caption{An example of simulated data points from the data-generating model, where the regression function is quadratic when $X\leq 0.1$ and linear otherwise.}\label{sim2}\end{figure}

We consider the data-generating process 
\begin{equation*}
    Y=
    \begin{cases}
      250(X+0.1)^2+\varepsilon, & \text{if}\ 0<X\leq 0.1, \\
      100X+\varepsilon, & \text{if}\ 0.1<X<1,
    \end{cases}
\end{equation*}
where the predictor $X$ is generated i.i.d.\  from  $U(0,1)$, and the random error $\varepsilon$'s are  i.i.d.\  from $N(0,1)$. 
It can be seen that the underlying true regression function is quadratic on the left and linear  on the right. Figure \ref{sim2} is an example of a random sample of this model.
Our candidate methods are the Nadaraya-Watson estimator \citep{nadaraya1964estimating} with Gaussian kernel and linear regression. 
The Nadaraya-Watson estimator is estimated by ``npreg" from the R package ``np" with bandwidth selected by 
least-squares cross-validation. 
Intuitively, the Nadaraya-Watson estimator may have better performance on the quadratic part and the linear regression
may have better performance on the linear part. 
We compare four kinds of TCVs with  weights:
\begin{eqnarray*}
 W_{n,0.5}(X) &= 
 \begin{cases}
 0.5,& X<0.1,\\
 0.5, & X\geq 0.1,
 \end{cases}
 \quad
  W_{n,0.8}(X) &= 
 \begin{cases}
 0.8,& X<0.1,\\
 0.2, & X\geq 0.1,
 \end{cases}\\
  W_{n,0.9}(X) &= 
 \begin{cases}
 0.9,& X<0.1,\\
 0.1, & X
 \geq 0.1,
 \end{cases}
 \quad
  W_{n,1}(X) &= 
 \begin{cases}
 1,& X<0.1,\\
 0, & X\geq 0.1.
 \end{cases}
 \end{eqnarray*}
 
 We randomly generate a dataset with the size $200$ and apply the TCVs with the different weight functions to the candidate methods with the training and the test sizes $n_1 = n_2 = 100$ and the number of splittings $K = 100$. The evaluation procedure is the same as in Section~\ref{s1}. The weighted MSE for local performance, performance outside the local region,  and the overall performance are considered. The MSEs and their standard errors based on $500$ replications for the candidate methods and TCVs are shown in Table~\ref{sim2_1_n_2}.
The notations ``NW" and ``Linear" stand for the Nadaraya-Watson estimator and linear regression, respectively. The notation ``$\text{TCV}_{\alpha}$" with $\alpha = 0.5,0.8,0.9,0.1$ stand for the TCVs with the corresponding weight $W_{n,\alpha}$, respectively.

\begin{table}[!ht]
\centering
\caption{MSEs from the Nadaraya-Watson estimator, linear regression, and TCVs with different weights. We consider  their local region performances, outside local region performances, and overall performances. The standard errors of the MSEs are shown in parentheses. }
\label{sim2_1_n_2}
\begin{tabular}{lllllll}\hline \noalign{\smallskip}
&NW    & Linear    & $\text{TCV}_{0.5}$     & $\text{TCV}_{0.8}$     & $\text{TCV}_{0.9}$     & $\text{TCV}_1$                  \\\noalign{\smallskip}\hline\noalign{\smallskip}
Local& 0.126     & 0.186     & 0.186     & 0.162     & 0.152     & 0.150      \\
      & (0.001) & (0.001) & (0.001) & (0.002) & (0.002) & (0.002) \\\noalign{\smallskip}
Outside& 1.120      & 0.927     & 0.927     & 1.013     & 1.043     & 1.051     \\
      & (0.003) & (0.001) & (0.001) & (0.005) & (0.005) & (0.005) \\\noalign{\smallskip}
Overall & 1.245     & 1.112     & 1.112     & 1.176     & 1.195     & 1.201     \\
      & (0.003) & (0.001) & (0.001) & (0.004) & (0.004) & (0.004)\\\noalign{\smallskip}\hline
\end{tabular}
\end{table}

We  see that the Nadaraya-Watson estimator performs better in the local region ($X<0.1$), and linear regression performs better outside the local region $(X>0.1$).
Additionally, linear regression has better overall performance. With $W_{n,0.5}$, the TCV is equivalent to CV, and it prefers
linear regression. As the weight in the local region increases, the performance of the TCV gets closer to that of the Nadaraya-Watson estimator, with improved local region performance.
Nevertheless, as a tradeoff, it has worse global and outside the local region performances. The results show that the TCV can address the problems where we need a balance between local and overall performance. 

\subsection{Simulation 3: a high-dimensional case}
In this example, we apply TCV in a high-dimensional setting. 
We consider the data-generating process 
\begin{equation*}
\begin{split}
Y&= 2\exp\bigl(-5{X^{(1)}}^2\bigr) + 2X^{(1)} + X^{(2)} + 0.5X^{(3)} + 0.1X^{(4)}
+\varepsilon.
\end{split}
\end{equation*}
The $1000$ predictors $X^{(1)},\dots,X^{(1000)}$ follow a multivariate normal distribution with mean $\bm{0}$ and covariance matrix $V\in \mathbb{R}^{1000}$ with $V_{i,j}=0.1^{|i-j|}$.
The random error $\varepsilon$'s are  i.i.d.\  from $N(0,1)$. We want to find out the best candidate method for the local region
\[
\{(x^{(1)}, x^{(2)}\cdots x^{(1000)})^{\T}:  (x^{(1)}, x^{(2)})^\T \in (-0.5,0.5)^2, (x^{(3)}\cdots x^{(1000)})^\T\in \mathbb R^{998}\}.
\]
Four candidate models/procedures are considered. They are random forest and lasso regression fitted on the complete dataset, and their local versions fitted on the local region data.

The nonlinear term $\exp\bigl(-5{X^{(1)}}^2\bigr)$ has a large influence on the regression function when $X^{(1)}$ is near 0. According to this, the random forest may have the best performance in the local region centered at 0. However, lasso regression also has a chance to overperform the random forest if the influence of the nonlinearity is relatively small compared with the advantage obtained from sparsity. The local candidate models are more specific to the local region compared with their global counterparts, but they have fewer data.

We randomly generate datasets with size 200, evaluate the candidate methods and CV selection results, and  independently repeat this process 100 times. The other simulation settings are the same as previous subsections. The random forest is fitted by 500 trees, and it selects 32 variables each time. The tuning parameter of the lasso regression is selected by the 10-fold CV. 

The MSEs and their standard errors  are shown in Table~\ref{sim3}.
The notations ``lasso" and ``RF" stand for lasso regression and random forest based on the complete dataset, respectively, and ``lasso\_local" and ``RF\_local" stand for the corresponding local versions. It can be seen that the random forest based on the complete dataset has the best local performance, and lasso regression based on the complete dataset has the best overall performance. 
 The TCV selects the complete dataset-based random forest for most of the times, and CV often selects the complete dataset-based lasso regression. Consequently, the TCV has better local performance and the CV has better overall performance, as expected.
\begin{table}[!ht]
\caption{MSEs from lasso regression and random forest built on all or local data, TCV, and CV. We consider  their local region performances and overall performances. The standard errors of the MSEs are shown in parentheses. The bold numbers stand for the better one out of the CV and TCV. A t-test at the significance level of 0.05 shows that there is no significant difference between the local performances of `lasso'(with MSE 2.02) and `lasso\_local'(with MSE 2.00).}
\label{sim3}
\centering
\begin{tabular}{lllllll}
\hline \noalign{\smallskip}
         & lasso  & RF & lasso\_local & RF\_local & TCV    & CV     
       \\\noalign{\smallskip}\hline\noalign{\smallskip}
Local    & 2.02   & 1.47           & 2.00       & 1.90                & $\bm{1.62}$   & 2.02   \\
         & (0.02) & (0.01)         & (0.02)     & (0.01)              & (0.03) & (0.02) \\\noalign{\smallskip}
Overall & 1.74   & 2.78           & 7.48       & 7.86                & 3.85   & $\bm{1.75}$   \\
         & (0.01) & (0.02)         & (0.13)     & (0.05)              & (0.23) & (0.02) \\\noalign{\smallskip}\hline
\end{tabular}
\end{table}
\subsection{Boston housing data}\label{boston}
We demonstrate the application of the TCV using the Boston housing data from \cite{harrison1978hedonic}
with 506 observations. 
They
fitted a model for the median value of owner-occupied homes based on the model
\begin{equation}
\label{hdnic}
\begin{split}
\log(MV)=&\ a_1+a_2RM^2+a_3AGE+a_4\log(DIS)+a_5\log(RAD)+\\
&\ a_6TAX+a_7PTRATIO+a_8B+a_9\log(LSTAT)+a_{10}CRIM+\\
&\ a_{11}ZN+a_{12}INDUS+a_{13}CHAS+a_{14}NOX^2+\varepsilon.
\end{split}
\end{equation} 

Instead of fitting a model of the overall house price like \eqref{hdnic}, we consider the house prices of relatively new buildings.  Since the age of a house is one of the most important factors that a home buyer typically considers, it will definitely make a difference in terms of the price. According to this fact, other candidate models may perform better than the global model \eqref{hdnic}.  

The predictor $AGE$ is the proportion of owner-occupied homes built prior to 1940, which measures the overall age of the houses in a census tract. Here, we are particularly interested in the medium home value in relatively new areas with  less than $50\%$ of the houses built before 1940 ($AGE<50$). There are 147 observations with $AGE<50$ in the data. 
We consider the following three natural candidate models: 
\begin{enumerate}
\item $\delta_1$ uses the model~\eqref{hdnic} with all the available data.
\item $\delta_2$ performs an additive regression, using the model
\begin{equation}
\label{hdnic2}
\begin{split}
\ \log(MV)=&\ a_1+a_2f(RM^2)+a_3f(AGE)+a_4f(\log(DIS))+\\
&\ a_5f(\log(RAD))+a_6f(TAX)+a_7f(PTRATIO)+\\
&\ a_8f(B)+\ a_9f(\log(LSTAT))+a_{10}f(CRIM)+\\
&\ a_{11}f(ZN)+ a_{12}f(INDUS)+a_{13}CHAS+\\
&\ a_{14}f(NOX^2)+\varepsilon,
\end{split}
\end{equation}
with all available data, where each $f$ represents a smoothing spline with  three degrees of freedom 
 \item $\delta_3$ is from model~\eqref{hdnic}  but only based on the local data with $AGE < 50$.
\end{enumerate}
 Note that the local version of $\delta_2$ is not considered as a candidate since we do not have enough local data to support the nonparametric estimation.

We first randomly set aside $20\%$ of the observations as the evaluation set and apply half-half splitting 100 times on the rest of the observations for the CV and TCV.
 To avoid not having enough local data in the randomly splitted data, we apply a stratified sampling.  The weight for the TCV is 
$W_{n}(X) = \mathbbm{1}(AGE<50)$.
We calculate the weighted MSE based on the local region and the overall MSE for the three candidate models together with the TCV and CV on the evaluation set. The above procedure is independently repeated 500 times. The MSEs, together with  their  standard errors, for the candidate models, the TCV, and  CV are shown in Table~\ref{bsrealdata}. 

\begin{table}[!ht]
\centering
\caption{MSEs of $\delta_1$, $\delta_2$, $\delta_3$, TCV, and  CV.  We consider both their local region performances and overall performances. The standard errors of the MSEs are shown in parentheses. The bold numbers stand for the better one out of the CV and TCV.}\label{bsrealdata}
\resizebox{\columnwidth}{!}{%
\begin{tabular}{llllll}\hline \noalign{\smallskip}
&$\delta_1$     & $\delta_2$         & $\delta_3$  & TCV         & CV                      \\\noalign{\smallskip}\hline \noalign{\smallskip}
 Local & 0.0049& 0.0037 & 0.0018 & $\bm{0.0018}$ & 0.0037 \\
          & $(6.7\times10^{-5})$ & $(6.4\times10^{-5})$    & $(2.4\times10^{-5})$ & $(2.5\times10^{-5})$    & $(6.4\times10^{-5})$    \\\noalign{\smallskip}
Overall    & 0.0360 & 0.0303 & 0.1710 & 0.1707 & $\bm{0.0304}$ \\
          & $(3.3\times10^{-4})$  & $(3.1\times10^{-4})$   & (0.007) & (0.007)   & $(3.1\times10^{-4})$   \\ \noalign{\smallskip}\hline
\end{tabular}
}
\end{table}
It can be seen that $\delta_3$  has the best performance in the local region where $AGE<50$. 
For the overall performance, $\delta_2$ is the best. The TCV always selected $\delta_3$ and the CV always selected $\delta_2$. 
Clearly, the TCV has excellent performance since it selected the candidate with the best performance
for the region of interest. 

\section{Concluding Remarks}\label{conclusion}
We have proposed the TCV for selecting the  best candidate regression procedure defined via  weighted $L_2$ losses. An application of the TCV is to find a candidate method with the best performance for a local region. With a proper data splitting, our method can consistently identify the best-performing candidate that possibly varies with the sample size. Simulation and real data examples have illustrated that the TCV can outperform the regular CV when the candidate methods do not rank uniformly over the local region and outside that region.

With the availability of large numbers of observations, often with high input dimensions, highly adaptive non-traditional methods can better approximate complicated regression functions. In this background, the traditional framework of a fixed best model or procedure waiting to be identified may be overly simplistic in real applications. In this paper, we utilize a triangular array setup to facilitate the needed flexibility for selecting the dynamically best candidate. This more dynamic framework may be generally helpful or even necessary in establishing adaptive learning theories for high-dimensional and big data when data splittings are involved. 

It is interesting to observe that when a weighting function is used to define the loss of interest, it may have a major impact on how we need to split the data. Indeed, when we care most about a small region, for instance, the task of finding out the best candidate procedure among close competitors in that region becomes harder, compared with that of identifying the globally best. Then, we need to shift the data splitting ratio more towards the evaluation part. Our main theorem gives sufficient conditions to enable consistent selection in terms of the natures of the weighting function and the convergence rates of the candidate procedures. We have also observed that the TCV may exhibit poor performance or high variability in terms of the global performance. Therefore, a proper CV method should be chosen according to the interest of the application. 

In this work, the number of candidate procedures to be compared is essentially not allowed to grow as the sample size increases. One interesting future direction is to handle the situation where the list of candidate procedures expands when more observations become available, which provides more flexibility in regression modeling. 

\section{Proof of the Main Theorem}\label{appendix_A}

Since the number of candidate methods in $\mathcal M$ is upper bounded, it suffices to show that the TCV is  $W_n$-consistent when comparing  the $(W_n, l_n,  c_{(n_1,n)})$-best candidate method $\delta_{\bn}$ and every other candidate method from $\mathcal M$. We take an arbitrary candidate method that is not $\delta_{\bn}$ from $\mathcal M$ and denote it by $\delta_{\wn}$.

In this proof, we will show that
when $l_n\leq n_1 < n$, the probability of selecting the $(W_n, l_n,  c_{(n_1,n)})$-better candidate 
\begin{equation}\label{goal}
P\left(TCV_{W_n}(\widehat{f}_{n_1}^{(\wn)})> TCV_{W_n}(\widehat{f}_{n_1}^{(\bn)})\right)\rightarrow 1,
\end{equation}
as $n\rightarrow\infty$.
 Since $\resn = f(\predn) + \errorn$, we have \begin{equation*}
\begin{split}
TCV_{W_n}(\widehat{f}_{n_1}^{(j)})
=& \sum^n_{i=n_1+1}\errorn^2\cdot W_n(\predn)+\sum^n_{i=n_1+1}(f(\predn)-\widehat{f}_{n_1}^{(j)}(\predn))^2\cdot \\
 &W_n(\predn)+2\sum^n_{i=n_1+1}\errorn(f(\predn)-\widehat{f}_{n_1}^{(j)}(\predn))
\cdot W_n(\predn),
\end{split}
\end{equation*}
where $j = 1$ or $2$.
We have
\begin{align}
	&TCV_{W_n}(\widehat{f}_{n_1}^{(\wn)})- TCV_{W_n}(\widehat{f}_{n_1}^{(\bn)}) \leq 0 \nonumber\\
\iff& 2\sum^n_{i=n_1+1}\errorn(\widehat{f}_{n_1}^{(\wn)}(\predn)-\widehat{f}_{n_1}^{(\bn)}(\predn) )\cdot W_n(\predn)\geq \mathcal K^{(\wn)} - \mathcal K^{(\bn)},\label{eq_TCV_ineq_equivalence}
\end{align}
where
\begin{align}
	\mathcal K^{(\bn)} &= \sum^n_{i=n_1+1}(f(\predn)-\widehat{f}_{n_1}^{(\bn)}(\predn))^2\cdot W_n(\predn),\\
	\mathcal K^{(\wn)} &= \sum^n_{i=n_1+1}(f(\predn)-\widehat{f}_{n_1}^{(\wn)}(\predn))^2\cdot W_n(\predn).
\end{align}
Denote the event 
$\{\mathcal K^{(\wn)} - \mathcal K^{(\bn)} > 0\}$
by $S_n$ and the training data by $\traindat$. 
Conditional on $\traindat$, the predictor data $\testX$ from the test set, and given $S_n$  holds, by the result from \eqref{eq_TCV_ineq_equivalence} and Chebyshev's inequality, we have
\begin{align}
&P\left(TCV_{W_n}(\widehat{f}_{n_1}^{(\wn)})\leq TCV_{W_n}(\widehat{f}_{n_1}^{(\bn)})|\traindat,\testX, S_n \right)\label{ineqq:1}\\
=& P\biggl(2\sum^n_{i=n_1+1}\epsilon_i\biggl(\widehat{f}_{n_1}^{(\wn)}(\predn)-\widehat{f}_{n_1}^{(\bn)}(\predn)\biggr)\cdot W_n(\predn) \geq \nonumber\\
&\mathcal K^{(\wn)} - \mathcal K^{(\bn)}|\traindat,\testX, S_n \biggr)\nonumber
\\
\leq&  \frac{Var(2\sum^n_{i=n_1+1}\epsilon_i(\widehat{f}_{n_1}^{(\wn)}(\predn)-\widehat{f}_{n_1}^{(\bn)}(\predn))\cdot W_n(\predn)|\traindat,\testX, S_n )}
{\left(\mathcal K^{(\wn)} - \mathcal K^{(\bn)}\right)^2}\nonumber\\
\leq&  \frac{4\sum^n_{n_1+1}(\widehat{f}_{n_1}^{(\wn)}(\predn)-\widehat{f}_{n_1}^{(\bn)}(\predn))^2\cdot (W_n(\predn))^2\cdot\overline\sigma^2}
{\left(\mathcal K^{(\wn)} - \mathcal K^{(\bn)}\right)^2} \leq  Q_n,\nonumber
\end{align}
almost surely, where
$Q_n = \frac{4\overline\sigma^2\sum^n_{n_1+1}(\widehat{f}_{n_1}^{(\wn)}(\predn)-\widehat{f}_{n_1}^{(\bn)}(\predn))^2\cdot W_n(\predn)\cdot S_{W_n}}
{\left(\mathcal K^{(\wn)} - \mathcal K^{(\bn)}\right)^2}$
(recall that  $S_{W_n}$ denotes $\text{ess-sup}_{\bm{x}}(W_n(\bm{x}))$ ).
Therefore, \eqref{ineqq:1} is upper bounded by
$\min\left(1, Q_n\right)$,
Thus, for the unconditional probability we have 
\begin{equation*}
\begin{split}
& P\left(TCV_{W_n}(\widehat{f}_{n_1}^{(\wn)})< TCV_{W_n}(\widehat{f}_{n_1}^{(\bn)})\right)\\
&=E\left[P\left(\{TCV_{W_n}(\widehat{f}_{n_1}^{(\wn)})< TCV_{W_n}(\widehat{f}_{n_1}^{(\bn)})\}\cap S_n|\traindat,\testX \right)\right]+\\
&\quad \, E\left[P\left(\{{TCV_{W_n}(\widehat{f}_{n_1}^{(\wn)})< TCV_{W_n}(\widehat{f}_{n_1}^{(\bn)}})\}\cap S_n^c|\traindat,\testX \right)\right]\\
&\leq E\left[P\left(\{TCV_{W_n}(\widehat{f}_{n_1}^{(\wn)})< TCV_{W_n}(\widehat{f}_{n_1}^{(\bn)})\}|\traindat,\testX,S_n \right)\right]+
E\left[P\left(S_n^c|\traindat,\testX \right)\right]\\
&\leq E\min(1,Q_n)+P(S_n^c).
\end{split}
\end{equation*} 
For the bound of $P(S_n^c)$, we first assume that for $l_n\leq n_1<n$, there exists an upper bounded positive sequence $\aln$, such that
\begin{equation}\label{eq:2}
P\left(\frac{\mathcal K^{(\wn)}}{\mathcal K^{(\bn)}}\geq 1+\aln\right)\rightarrow 1,
\end{equation}
as $n\rightarrow\infty$.
The above inequality implies that $P(S_n)\rightarrow 1$ as $n\rightarrow\infty$. For the bound of $E\min(1,Q_n)$, we have
\begin{equation*}
\begin{split}
&P\left(\frac{\mathcal K^{(\bn)}}{\mathcal K^{(\wn)}}\leq \frac{1}{1+\aln}\right)\\
&= P\left(\frac{\mathcal K^{(\wn)}-\mathcal K^{(\bn)}}{\mathcal K^{(\wn)}}\geq1-\frac{1}{1+\aln}\right)\\
&\leq  P\left(\left( \frac{\mathcal K^{(\wn)}-\mathcal K^{(\bn)}}
{(1-1/(1+\aln))\mathcal K^{(\wn)}}\right)^2 \geq 1\right) \\
&=  P\left(\left( \frac{\mathcal K^{(\wn)}-\mathcal K^{(\bn)}}
{(1-1/(1+\aln))\mathcal K^{(\wn)}}\right)^2 \cdot Q_n\geq  Q_n\right)\\
&= P\left(Q_n\leq\frac{4\overline\sigma^2\sum^n_{n_1+1}\left(\widehat{f}_{n_1}^{(\wn)}(\predn)-\widehat{f}_{n_1}^{(\bn)}(\predn)\right)^2W_n(\predn)\cdot S_{W_n}}{\left((1-1/(1+\aln))\mathcal K^{(\wn)}\right)^2}\right).
\end{split}
\end{equation*}
Combining the above result and inequality~\eqref{eq:2}, we have
\begin{equation}\label{An}
P\left(Q_n\leq\frac{4\overline\sigma^2\sum^n_{n_1+1}\left(\widehat{f}_{n_1}^{(\wn)}(\predn)-\widehat{f}_{n_1}^{(\bn)}(\predn)\right)^2W_n(\predn)\cdot S_{W_n}}{\left((1-1/(1+\aln))\mathcal K^{(\wn)}\right)^2}\right)
\rightarrow 1,	
\end{equation}
as $n\rightarrow\infty$.  By the fact that $(a + b)^2\leq 2(a^2 + b^2)$, we have
\begin{equation}\label{eq:1}
\sum^n_{n_1+1}\left(\widehat{f}_{n_1}^{(\wn)}(\predn)-\widehat{f}_{n_1}^{(\bn)}(\predn)\right)^2\cdot W_n(\predn)\leq 2\bigl(\mathcal K^{(\wn)}+\mathcal K^{(\bn)}\bigr).
\end{equation}
Given \eqref{eq:2}, by combining \eqref{An} and \eqref{eq:1}, we obtain
 \begin{equation}\label{eq:3}
P\left(Q_n\leq \frac{8\overline\sigma^2(\mathcal K^{(\wn)}+\mathcal K^{(\bn)})\cdot S_{W_n}}{\left((1-1/(1+\aln))\mathcal K^{(\wn)}\right)^2}\right)\rightarrow1,
\end{equation}
as $n\rightarrow\infty$.
When \eqref{eq:2} holds, we also have
\begin{align}
	&P\left(\frac{\mathcal K^{(\wn)}+\mathcal K^{(\bn)}}{\mathcal K^{(\wn)}}\leq 1+\frac{1}{1+\aln}\right)\nonumber\\
	&=P\left(\frac{8\overline\sigma^2(\mathcal K^{(\wn)}+\mathcal K^{(\bn)})\cdot S_{W_n}}{\left((1-1/(1+\aln))\mathcal K^{(\wn)}\right)^2}\leq \frac{8\overline\sigma^2(1+1/(1+\aln))\cdot S_{W_n}}{(1-1/(1+\aln))^2\mathcal K^{(\wn)}}\right)  \rightarrow1, \label{Bn}
\end{align}
as $n\rightarrow\infty$.
It follows from \eqref{eq:3} and \eqref{Bn} that
\begin{equation}\label{q28}
P\left(Q_n \leq\frac{8\overline\sigma^2(1+1/(1+\aln))\cdot S_{W_n}}{(1-1/(1+\aln))^2\mathcal K^{(\wn)}}
\right)\rightarrow1,
\end{equation}
as $n\rightarrow\infty$.
Therefore, by the fact that $Q_n \geq 0$ and the dominated convergence theorem, to control the upper bound of $E\min(1,Q_n)$, it suffices to show
\begin{equation}\label{eq:10}
\aln^2\cdot \mathcal K^{(\wn)}\big/S_{W_n}\overset{p}{\rightarrow}\infty .
\end{equation}

According to the above results, 
to prove Thereom~\ref{thm}, it suffices to show 
\eqref{eq:2} and \eqref{eq:10}.
Next, we will establish one inequality between $\mathcal K^{(\bn)}$ 
and $\| f-\widehat{f}_{n_1}^{(\wn)}\|^2_{2,W_n} $ and another inequality between $\mathcal K^{(\wn)}$ 
and  $\| f-\widehat{f}_{n_1}^{(\wn)}\|^2_{2,W_n}$.
We first derive the inequality between $\mathcal K^{(\bn)}$ 
and $\| f-\widehat{f}_{n_1}^{(\wn)}\|^2_{2,W_n} $.
%
We define $\mathcal D_{n,i,j}\triangleq(f(\predn)-\widehat f_{n_1}^{(j)}(\predn))^2\cdot W_n(\predn)-\| f-\widehat f_{n_1}^{(j)}\|_{2,W_n}^2.$
Let $E_{\traindat}(\cdot)$ and $Var_{\traindat}(\cdot)$ denote the expectation and variance conditional on $\traindat$, respectively. We have
\begin{align}
Var_{\traindat}(\mathcal D_{n,n_1+1,j})
&\leq E_{\traindat}\left((f(\prednone)-\widehat f_{n_1,j}(\prednone))^4\cdot\left(W_n(\prednone)\right)^2\right)\nonumber\\
&\leq \| f-\widehat f_{n_1}^{(j)}\|_{4,W_n}^4\cdot S_{W_n}. \label{varineq}
\end{align}
The above inequality holds for both  the $(W_n, l_n,   c_{(n_1,n)})$-better candidate $\delta_{\wn}$ and worse candidate $\delta_{\bn}$. 
Therefore, by \eqref{varineq} and Chebyshev's inequality, we obtain that  for each $a>0$,
\begin{equation*}
\begin{split}
P_{\traindat}\left(\mathcal K^{(\bn)}-n_2\| f-\widehat{f}_{n_1}^{(\bn)}\|^2_{2,W_n}\geq a   \right)
&\leq  \frac{n_2Var_{\traindat}(\mathcal D_{n, n_1+1, \bn})}{a^2}\\
&\leq  \frac{n_2\| f-\widehat f_{n_1}^{(\bn)}\|_{4,W_n}^4 \cdot S_{W_n}}{a^2},
\end{split}
\end{equation*}
where $P_{\traindat}$ is the probability conditional on the training data $\traindat$.
We take $a=\bet_{1,n_1} n_2\| f-\widehat{f}_{n_1}^{(\bn)}\|^2_{2,W_n}$, where $\bet_{1,n_1}>0$ and will be determined latter. 
According to the above inequality, we have
\begin{align}
P_{\traindat}\left( \mathcal K^{(\bn)}\geq(1+\bet_{1,n_1})n_2\| f-\widehat{f}_{n_1}^{(\bn)}\|^2_{2,W_n}   \right)
\leq  \frac{\| f-\widehat f_{n_1}^{(\bn)}\|_{4,W_n}^4\cdot S_{W_n}}{ \bet_{1,n_1}^2n_2\| f-\widehat{f}_{n_1}^{(\bn)}\|^4_{2,W_n}  } \label{ber}.
\end{align}
We denote event $\left\{   \| f-\widehat{f}_{n_1}^{(\wn)}\|^2_{2,W_n} /  \| f-\widehat{f}_{n_1}^{(\bn)}\|^2_{2,W_n} \geq 1 + c_{(n_1,n)} \right\}$ by $D_n$. 
Then, according to Definition~\ref{def_bestdef},  we have 
\begin{equation}\label{Dncon}
P\left( D_n\right)\rightarrow 1,
\end{equation}
as $n\rightarrow\infty$.
Without losing generality, we require that the sequence $\{c_{(n_1,n)}\}$ is upper bounded (otherwise, we can always replace the original $\{c_{(n_1,n)}\}$ by a bounded sequence). 
To obtain the inequality between $\mathcal K^{(\bn)}$ 
and  $\| f-\widehat{f}_{n_1}^{(\wn)}\|^2_{2,W_n} $, we take
$	
\bet_{1,n_1}=\| f-\widehat{f}_{n_1}^{(\wn)}\|^2_{2,W_n}/\left((1+c_{(n_1,n)}/2\right) \| f-\widehat{f}_{n_1}^{(\bn)}\|^2_{2,W_n})-1.
$
Conditional on $D_n$, we have
\begin{align}
\bet_{1,n_1} & = \frac{\| f-\widehat{f}_{n_1}^{(\wn)}\|^2_{2,W_n} - \| f-\widehat{f}_{n_1}^{(\bn)}\|^2_{2,W_n}(1+c_{(n_1,n)}/2)}{\| f-\widehat{f}_{n_1}^{(\bn)}\|^2_{2,W_n}(1+c_{(n_1,n)}/2)} \nonumber\\
&\geq \frac{\| f-\widehat{f}_{n_1}^{(\wn)}\|^2_{2,W_n} - \| f-\widehat{f}_{n_1}^{(\wn)}\|^2_{2,W_n}\frac{1+c_{(n_1,n)}/2}{1+c_{(n_1,n)}}}{\| f-\widehat{f}_{n_1}^{(\bn)}\|^2_{2,W_n}(1+c_{(n_1,n)}/2)}\nonumber\\
&= \frac{c_{(n_1,n)}\| f-\widehat{f}_{n_1}^{(\wn)}\|^2_{2,W_n}}
{2(1+c_{(n_1,n)})(1+c_{(n_1,n)}/2)\| f-\widehat{f}_{n_1}^{(\bn)}\|^2_{2,W_n}}. \label{eq:4}
\end{align}
According to \eqref{eq:4} and \eqref{ber}, we obtain
\begin{align}
&P\left(  \mathcal K^{(\bn)} \geq \frac{n_2}{1+c_{(n_1,n)}/2}\| f-\widehat{f}_{n_1}^{(\wn)}\|^2_{2,W_n}  \right) \nonumber \\
\leq & P\left(\left\{ \mathcal K^{(\bn)}\geq \frac{n_2}{1+c_{(n_1,n)}/2}\| f-\widehat{f}_{n_1}^{(\wn)}\|^2_{2,W_n} \right\} \cap D_n \right) + P(D_n^c),\nonumber \\
\leq & P\left( \mathcal K^{(\bn)}\geq \frac{n_2}{1+c_{(n_1,n)}/2}\| f-\widehat{f}_{n_1}^{(\wn)}\|^2_{2,W_n}  \mid D_n \right) + P(D_n^c),\nonumber \\
=& E\biggl(P_{\traindat}\left(\mathcal K^{(\bn)} \geq(1+\bet_{1,n_1})n_2\| f-\widehat{f}_{n_1}^{(\bn)}\|^2_{2,W_n} \right)\mid D_n \biggr)+ P(D_n^c)\nonumber\\
\leq &  E(Q_n^{(1)})+ P(D_n^c), \label{ber2}
\end{align}
where
$
Q_n^{(1)} \triangleq \min\biggl\{1, \frac{4(1+c_{(n_1,n)})^2(1+c_{(n_1,n)}/2)^2}{c_{(n_1,n)}^2}\cdot \frac{\| f-\widehat f_{n_1}^{(\bn)} \|_{4,W_n}^4 \cdot S_{W_n}}{ n_2\| f-\widehat{f}_{n_1}^{(\wn)}\|^4_{2,W_n}  } \biggr\}.
$

Next, we derive the other inequality that compares
$\mathcal K^{(\wn)}$ 
and \\ $\| f-\widehat{f}_{n_1}^{(\wn)}\|^2_{2,W_n} $.
Let $\{\bet_{2, n_1}\}$ be a positive sequence whose complete definition will be given latter. We require it to be bounded above by $1$.
By Chebyshev's inequality and \eqref{varineq}, we have
\begin{align}
&P(\mathcal K^{(\wn)}  <(1-\bet_{2, n_1} )n_2\| f-\widehat{f}_{n_1}^{(\wn)}\|^2_{2,W_n})	 \nonumber \\
&=P(-\mathcal K^{(\wn)} + n_2\| f-\widehat{f}_{n_1}^{(\wn)}\|^2_{2,W_n} >\bet_{2, n_1} n_2\| f-\widehat{f}_{n_1}^{(\wn)}\|^2_{2,W_n})	 \nonumber \\
&=  E\biggl(P_{\traindat}\biggl(-\mathcal K^{(\wn)}+ n_2\| f-\widehat{f}_{n_1}^{(\wn)}\|^2_{2,W_n}  >\bet_{2, n_1} n_2\| f-\widehat{f}_{n_1}^{(\wn)}\|^2_{2,W_n}\biggr)\biggr) \nonumber \\
&\leq  E(Q_n^{(2)}), \label{ineq2}
\end{align}
where
$
Q_n^{(2)} \triangleq \min \biggl\{1,\frac{\| f-\widehat f_{n_1}^{(\wn)}\|_{4,W_n}^4\cdot S_{W_n}}{\bet_{2, n_1}^2 n_2\| f-\widehat{f}_{n_1}^{(\wn)}\|^4_{2,W_n}}\biggr\}.
$

To show \eqref{eq:2}, we take $\bet_{2, n_1} = 1-\frac{1+ c_{(n_1,n)}/4 }{1+c_{(n_1,n)}/2}$ and $\aln = c_{(n_1,n)}/4$. By combining \eqref{ber2} and \eqref{ineq2}, we have
$
P\left( \mathcal K^{(\wn)}\geq (1+\aln)\cdot\mathcal K^{(\bn)} \right)\geq  1-P(D_n^c) -E(Q_n^{(1)})-E(Q_n^{(3)}),$
where $Q_n^{(3)}$ is $Q_n^{(2)}$ with $\bet_{2, n_1} = 1-\frac{1+ c_{(n_1,n)}/4 }{1+c_{(n_1,n)}/2}$.
Recall that the sequence $\{c_{(n_1,n)}\}$ is upper bounded. Therefore,  by the dominated convergence theorem, it remains to show that	
\begin{align}
\frac{1}{c_{(n_1,n)}^2}\cdot\frac{\| f-\widehat f_{n_1}^{(\bn)}\|_{4,W_n}^4\cdot S_{W_n} }{ n_2\| f-\widehat{f}_{n_1}^{(\wn)}\|^4_{2,W_n} }&< 	\frac{1}{c_{(n_1,n)}^2}\cdot\frac{\| f-\widehat f_{n_1}^{(\bn)}\|_{4,W_n}^4\cdot S_{W_n} }{ n_2\| f-\widehat{f}_{n_1}^{(\bn)}\|^4_{2,W_n} }\rightarrow 0,\nonumber \\
	\frac{1}{c_{(n_1,n)}^2}\cdot \frac{\| f-\widehat f_{n_1}^{(\wn)}\|_{4,W_n}^4\cdot S_{W_n}}{ n_2\| f-\widehat{f}_{n_1}^{(\wn)}\|^4_{2,W_n}}&\rightarrow 0, \label{eq_bn_bn_0}
\end{align} 
as $n\rightarrow\infty$ and $l_n\leq n_1<n$. According to Condition~\ref{cond4}, to obtain the above results, it suffices to show that
$
 M_{n}^4\cdot S_{W_n}/(c_{(n_1,n)}^2\cdot n_2)\rightarrow 0,
$
as $n\rightarrow\infty$, and this is required by Theorem~\ref{thm}.

For \eqref{eq:10}, recall that we have $\aln = c_{(n_1,n)}/4$. Therefore, it suffices to show that 
\begin{equation}\label{eq_toproof2}
c_{(n_1,n)}^2\cdot \mathcal K^{(\wn)}\big/S_{W_n}\overset{p}{\rightarrow}\infty,
\end{equation}
as $n\rightarrow\infty$. By \eqref{ineq2}, \eqref{eq_bn_bn_0}, and the fact that $1 - \bet_{2, n_1} = \frac{1+ c_{(n_1,n)}/4 }{1+c_{(n_1,n)}/2}$, which is bounded between $1/2$ and 1, we have that the rate of $K^{(\wn)}$ is lower bounded by $q_{(n_1,n)}$.
Therefore, we obtain \eqref{eq_toproof2} by the requirement \textbf{(ii)} from Theorem~\ref{thm}. Thus, we obtain \eqref{eq:2} and \eqref{eq:10} and complete the proof.

\section{Proof of Proposition~\ref{prop_1}}\label{Appendix_prop1Proof}
By the fact that  $(\sqrt{1 + n_1^{-1/3}} - 1)/n_1^{-1/3}\rightarrow1/2$ as $n_1\rightarrow\infty$, we have that when $n$ is sufficiently large and $n_1\in [l_n,n)$, 
\begin{align*}
&\| f-\widehat{f}_{n_1}^{(\wn)} \|_{2,W_n}^2 \geq (1 + n_1^{-1/3})\| f- \widehat{f}_{n_1}^{(\bn)}\|_{2,W_n}^2\\
\Rightarrow&
\| f-\widehat{f}_{n_1}^{(\wn)} \|_{2,W_n} \geq \bigl(1 + n_1^{-1/3}/3\bigr)\| f- \widehat{f}_{n_1}^{(\bn)}\|_{2,W_n}.		
\end{align*}
Therefore, it suffices to show that as $n\rightarrow\infty$,
\begin{align}
	\text{when }n\in\mathbb S_1,\ & P\Biggl(n_1^{-1/3}\cdot \frac{\|f - \hat f^{(1)}_{n_1}\|_2^2}{\|f - \hat f^{(2)}_{n_1}\|_2^2-\|f - \hat f^{(1)}_{n_1}\|_2^2} < 1\Biggr)\rightarrow1, \label{eq_cResult1}
\\	
\text{when }n\in\mathbb S_2,\ & P\Biggl(n_1^{-1/3}\cdot \frac{\|f - \hat f^{(2)}_{n_1}\|_2^2}{\|f - \hat f^{(1)}_{n_1}\|_2^2-\|f - \hat f^{(2)}_{n_1}\|_2^2} < 1\Biggr)\rightarrow1.\label{eq_cResult2}
\end{align}

Note that for $m = 1,2$, 
\begin{align} 
\|f - \hat f^{(m)}_{n_1}\|_2^2 =& \sum_{j=1}^{p^{(m)}_{n_1}}(\beta_j - \hat \beta_j^{(n_1)})^2 + \sum_{j=p^{(m)}_{n_1}+1}^{\infty}\beta_j^2,\label{eq_L1}\\
	\|f - \hat f^{(1)}_{n_1}\|_2^2 - \|f - \hat f^{(2)}_{n_1}\|_2^2 
	=& -\bigl(\beta_{p^{(2)}_{n_1}} - \hat \beta_{p^{(2)}_{n_1}}^{(n_1)}\bigr)^2 + \beta_{p^{(2)}_{n_1}}^2.\nonumber
	\end{align}

For $n\in\mathbb S_1$, since 
$p^{(2)}_{n_1} = n_1^{1/4}\in [n^{1/12}, n^{1/4}]$, we have
$ \beta_{p^{(2)}_{n_1}}^2 = 0$. Thus,
\begin{align*}
&\|f - \hat f^{(2)}_{n_1}\|_2^2 - \|f - \hat f^{(1)}_{n_1}\|_2^2 \\
	=& \biggl(\beta_{p^{(2)}_{n_1}} - \hat \beta_{p^{(2)}_{n_1}}^{(n_1)}\biggr)^2\\
	=& \biggl(\beta_{p^{(2)}_{n_1}} - \frac{1}{n_1}\sum_{i=1}^{n_1}f(x_i)\phi_{p^{(2)}_{n_1}}(x_i) - \frac{1}{n_1}\sum_{i=1}^{n_1}\epsilon_i\phi_{p^{(2)}_{n_1}}(x_i)\biggr)^2.
\end{align*}
Since $\beta_j$, $f(x_i)$, and $\phi_j(x_i)$ are all uniformly upper bounded by fixed constants for all $i$ and $j$, we have that $E(\beta_j -f(x)\phi_j(x) - \epsilon\phi_j(x))^2$ and $E(\beta_j -f(x)\phi_j(x) - \epsilon\phi_j(x))^3$ are uniformly upper bounded. Also, by
\begin{align*}
    E(\beta_j -f(x)\phi_j(x) - \epsilon\phi_j(x))^2 =& E(\beta_j -f(x)\phi_j(x))^2 + E \epsilon^2\phi_j(x)^2 - \\
    &2E \epsilon\phi_j(x)(\beta_j -f(x)\phi_j(x))\\
    =& E(\beta_j -f(x)\phi_j(x))^2 + E \epsilon^2\phi_j(x)^2 -0 \\
    \geq& E \epsilon^2\phi_j(x)^2\\
    =& 1,
\end{align*}
we have that $E(\beta_j -f(x)\phi_j(x) - \epsilon\phi_j(x))^2$ is uniformly  lower bounded away from 0. 
Therefore, by the Lyapunov CLT, we have
\begin{equation}\label{eq_LyapunovCLT}
	\frac{1}{\mathcal V_{p^{(2)}_{n_1}}}\cdot n_1\cdot \biggl(\beta_{p^{(2)}_{n_1}} - \hat \beta_{p^{(2)}_{n_1}}^{(n_1)}\biggr)\rightarrow_d N(0,1)
\end{equation}
as $n\rightarrow\infty$, where $\mathcal V_{p^{(2)}_{n_1}} = \sqrt{\sum_{i=1}^{n_1}E(\beta_{p^{(2)}_{n_1}} -f(x_i)\phi_{p^{(2)}_{n_1}}(x_i) - \epsilon\phi_{p^{(2)}_{n_1}}(x_i))^2}$.
  Next, we denote the uniform upper bound of $E(\beta_j -f(x)\phi_j(x) - \epsilon\phi_j(x))^2$ by $\mathcal C$. Then, 
\begin{align*}
 E\bigl((\beta_j - \hat \beta_j^{(n_1)})^2\bigr)
 =& \frac{1}{n_1^2}E\left(\left(\sum_{i=1}^{n_1}(\beta_{j} -f(x_i)\phi_{j}(x_i) - \epsilon\phi_{j}(x_i))\right)^2\right)\\
 =&\frac{1}{n_1^2}E\left(\sum_{i=1}^{n_1}\left(\beta_{j} -f(x_i)\phi_{j}(x_i) - \epsilon\phi_{j}(x_i)\right)^2\right)\leq\mathcal C/n_1.
\end{align*}
By Markov's inequality, for each $t>0$,
\begin{equation}\label{eq_markov_bound}
	 P\Biggl(\sum_{j=1}^{p^{(1)}_{n_1}}(\beta_j - \hat \beta_j^{(n_1)})^2 > t\Biggr)
 \leq  \frac{E\biggl(\sum_{j=1}^{p^{(1)}_{n_1}}(\beta_j - \hat \beta_j^{(n_1)})^2 \biggr)}{t}\leq  \frac{p^{(1)}_{n_1}\mathcal C}{n_1t}.
\end{equation}
Thus, 
\begin{equation}\label{eq_lossPart1}
	\sum_{j=1}^{p^{(1)}_{n_1}}(\beta_j - \hat \beta_j^{(n_1)})^2 = O_p\bigl(n_1^{-3/4}\bigr).
\end{equation}
Also, by $p^{(2)}_{n_1} = p^{(1)}_{n_1}+1\in [n^{1/12}, n^{1/4}]$, we have that (note that when $n\in\mathbb S_1\cup\mathbb S_2$, $n^{1/4}$ is an integer)
$
\sum_{j=p^{(1)}_{n_1}+1}^{\infty}\beta_j^2= 0 + \sum_{j = n^{1/4}}^\infty \beta_j^2 \leq \int_{n^{1/4}-1}^\infty \frac{1}{j^4} dj = \frac{1}{3}\cdot \frac{1}{(n^{1/4}-1)^3}.
$
Therefore, 
\begin{equation}\label{eq_betaSquare_rate}
	\sum_{j=p^{(1)}_{n_1}+1}^{\infty}\beta_j^2 = O_p(n^{-3/4}).
\end{equation}
According to \eqref{eq_lossPart1}, \eqref{eq_betaSquare_rate}, and the fact that $n_1<n$, we have that 
\begin{equation}\label{eq_L22_rate}
	\|f - \hat f^{(1)}_{n_1}\|_2^2 = O_p(n_1^{-3/4}). 
\end{equation}
Therefore, by \eqref{eq_LyapunovCLT}, \eqref{eq_L22_rate}, and the fact that $n_1^{-3/4}\cdot n_1 = n_1^{1/4}$, which goes to infinity at a slower rate than $n_1^{1/3}$, we obtain \eqref{eq_cResult1}.

For $n\in\mathbb S_2$,
we have
$
\|f - \hat f^{(1)}_{n_1}\|_2^2 - \|f - \hat f^{(2)}_{n_1}\|_2^2 
=   -\biggl(\beta_{p^{(2)}_{n_1}} - \hat \beta_{p^{(2)}_{n_1}}^{(n_1)}\biggr)^2 + \beta_{p^{(2)}_{n_1}}^2
=  -\biggl(\beta_{p^{(2)}_{n_1}} - \hat \beta_{p^{(2)}_{n_1}}^{(n_1)}\biggr)^2  + n_1^{-1/2}.
$
According to the result from \eqref{eq_LyapunovCLT}, $\bigl(\beta_{p^{(2)}_{n_1}} - \hat \beta_{p^{(2)}_{n_1}}^{(n_1)}\bigr)^2 $ goes to 0 at the exact rate $1/n_1$, which is faster than that of $n_1^{-1/2}$. Therefore, $\|f - \hat f^{(1)}_{n_1}\|_2^2 - \|f - \hat f^{(2)}_{n_1}\|_2^2 \rightarrow 0$ as $n\rightarrow\infty$ with the rate lower bounded by $1/n_1$. 
With a similar derivation of  \eqref{eq_lossPart1}, we have that 
\begin{equation}\label{eq_lossPart1_2}
	\sum_{j=1}^{p^{(2)}_{n_1}}(\beta_j - \hat \beta_j^{(n_1)})^2 = O_p\bigl(n_1^{-3/4}\bigr).
\end{equation}
We also have
$
\sum_{j=p^{(2)}_{n_1}+1}^{\infty}\beta_j^2= \sum_{j = n_1^{1/4}+1}^\infty \beta_j^2 \leq \int_{n_1^{1/4}}^\infty \frac{1}{j^4} dj = \frac{1}{3}\cdot n_1^{-3/4}.
$
Combining the above result with \eqref{eq_lossPart1_2}, we that
$
\|f - \hat f^{(2)}_{n_1}\|_2^2 = O_p(n_1^{-3/4}).
$
Therefore, according to the above results, we also obtain \eqref{eq_cResult2} and complete the proof.

\section{Proof of Proposition~\ref{prop_2}}\label{Appendix_prop2Proof}
First, we show the existence of the $n_1$ such that $n_2/n_1^{17/12}\rightarrow\infty$ as $n\rightarrow\infty$. Since $n_2/n_1^{17/12}$ = $(n - n_1)/n_1^{17/12}$, it suffices to have $n/n_1^{17/12}\rightarrow\infty$ as $n\rightarrow\infty$. According to $n^{12/17}/l_n = n^{19/51}$ and $n_1\in [l_n,n)$, we have verified the existence of such an $n_1$.

Next, we verify the requirement \textbf{(ii)}.
It can be seen from the derivation from Appendix~\ref{Appendix_prop1Proof} that  both $\|f - \hat f^{(1)}_{n_1}\|_2^2$ and $\|f - \hat f^{(2)}_{n_1}\|_2^2$ converge at the exact rate $n_1^{-3/4}$. Thus, $q_{(n_1,n)}^2 = n_1^{-3/4}$. Since $c_{(n_1,n)}^2 = n_1^{-2/3}/9$, when $n_2\cdot n_1^{-3/4}\cdot n_1^{-2/3} = n_2/n_1^{17/12}\rightarrow\infty$, the requirement \textbf{(ii)} is satisfied.

For the requirement \textbf{(i)},
it remains to show that $M_n =O_p(1).$
 According to the proof of Theorem 1 and Corollary 3.1. from \cite{zhang2015cross}, it suffices to show that as $p\rightarrow\infty$, 
$\bigl\|\sum_{j=p}^{\infty}\beta_j	\phi_{j}(x)\bigr\|_4^4 = O\bigl(\bigl\|\sum_{j=p}^{\infty}\beta_j	\phi_{j}(x)\bigr\|_2^4\bigr).$
First, we have
$$
\bigl\|\sum_{j=p}^{\infty}\beta_j	\phi_{j}(x)\bigr\|_2^4 
= 	\bigl(\sum_{j=p}^{\infty}\beta_j^2\bigr)^2
=\sum_{j=p}^{\infty}\beta_j^4 + 2\sum_{p\leq j<k}\beta_j^2\beta_k^2.$$ Second, for $a\leq b\leq c\leq d$ from the set $\{4\pi x, 4^2\pi x, 4^3\pi x,\cdots\}$, we have
$
\sin(a)\sin(b)\sin(c)\sin(d)	
= \frac{1}{8}(
\cos(a-b+c-d) + \cos(a-b-c+d)-
\cos(a-b+c+d) - \cos(a-b-c-d)	-
\cos(a+b+c-d) - \cos(a+b-c+d)+
\cos(a+b+c+d) + \cos(a+b-c-d)).
$
It can be seen that $a-b+c+d$, $a-b-c-d$, $a+b+c-d$, $a+b-c+d$, $a+b+c+d$ are all in the form of $\pi x$ times a non-zero even factor. Therefore, 
$
	\int_0^1(\sin(a)\sin(b)\sin(c)\sin(d))	
dx = \frac{1}{8}\int_0^1 (\cos(a-b+c-d) + \cos(a-b-c+d)+
 \cos(a+b-c-d)) dx.
$
It can been seen that for cases with $a<b<c<d$,  
$a = b<c<d$, $a < b=c<d$,  $a < b<c=d$, 
$a = b=c<d$, $a < b=c=d$, we have $a-b+c-d$, $a-b-c+d$,  and $a+b-c-d$ are all  in the form of $\pi x$ times a non-zero even factor. Therefore, $\int_0^1(\sin(a)\sin(b)\sin(c)\sin(d))dx	=0$ in these cases. For $a=b<c=d$,  we have
$
\int_0^1(\sin(a)\sin(b)\sin(c)\sin(d))	 dx=\frac{1}{8}(1 + 1 + 0 ) = \frac{1}{4}.
$
For $a=b=c=d$,  we have
$
\int_0^1(\sin(a)\sin(b)\sin(c)\sin(d))	 dx=\frac{1}{8}(1 + 1 + 1) = \frac{3}{8}.
$
Therefore, 
\begin{align*}
\Biggl\|\sum_{j=p}^{\infty}\beta_j	\phi_{j}(x)\Biggr\|_4^4
=&\int_0^1\Biggl(\sum_{j=p}^\infty 	(\beta_j	\phi_{j}(x))^4 + 6\sum_{p\leq j<k} 	(\beta_j	\phi_{j}(x))^2(\beta_k	\phi_{k}(x))^2\Biggr)dx\\
=&\frac{3}{8}\sum_{j=p}^{\infty}\beta_j^4 + \frac{3}{2}\sum_{p\leq j<k}\beta_j^2\beta_k^2.
\end{align*}
We have
$
\frac{\|\sum_{j=p}^{\infty}\beta_j	\phi_{j}(x)\|_4^4 }{\|\sum_{j=p}^{\infty}\beta_j	\phi_{j}(x)\|_2^4}
= \frac{3}{8}+ \frac{\frac{3}{4} \sum_{p\leq j<k}\beta_j^2\beta_k^2}{\sum_{j=p}^{\infty}\beta_j^4 + 2\sum_{i<k}\beta_j^2\beta_k^2} \leq \frac{3}{8} + \frac{3}{8}  =\frac{3}{4}.
$
Thus, we obtain $M_n =O_p(1)$ and complete the proof.

\section{Proof of Corollary~\ref{cor_MTCV_consistency}}
For the multiple splitting TCV by voting, we have
\begin{align*}
E\bigl(MTCV_{W_n}^v(\delta_{\bn})\bigr)
=& 
\frac{1}{K}\sum^{K}_{k=1}E(\mathbbm{1}(j = \arg\min_{i\in\mathcal J}TCV_{W_n,k}( \widehat{f}_{n_1,k}^{(i)})))\\
=&
P\left(TCV_{W_n}(\widehat{f}_{n_1}^{(\bn)})=  \min_{i\in\mathcal J}TCV_{W_n}(\widehat{f}_{n_1}^{(i)})\right)
\label{multisplit_v}	
\end{align*}
Therefore, when the requirements in Theorem~\ref{thm} are met, we have
$$
E\bigl(MTCV_{W_n}^v(\delta_{\bn})\bigr)\rightarrow 1,
$$
as $n\rightarrow\infty$. Since $MTCV_{W_n}^v(\delta_{\bn})\leq 1$, we have that
$
MTCV_{W_n}^v(\delta_{\bn})\rightarrow_p 1,
$
which implies that the multiple splitting TCV by voting is $W_n$-consistent.

For the multiple splitting TCV by averaging, we have
\begin{align*}
	&P\biggl(MTCV_{W_n}^a(\delta_{\bn}) =  \min_{i\in\mathcal J} MTCV_{W_n}^a(\delta_{i})\biggr)\\
	\geq& 1 - K\cdot P\left(TCV_{W_n}(\widehat{f}_{n_1}^{(\bn)})\geq  \min_{i\in\mathcal J_b}TCV_{W_n}(\widehat{f}_{n_1}^{(i)})\right).
\end{align*}
Since $K$ is upper bounded by a fixed constant, we also have the \\$W_n$-consistency of the multiple splitting TCV by averaging given the requirements in Theorem~\ref{thm}.

\section*{Acknowledgement}

This paper is based upon work supported by the National Science Foundation under grant number ECCS-2038603.

\bibliographystyle{apalike}

\bibliography{bibliography}

\begin{thebibliography}{}

\bibitem[Allen, 1974]{allen1974relationship}
Allen, D.~M. (1974).
\newblock The relationship between variable selection and data agumentation and
  a method for prediction.
\newblock {\em Technometrics}, 16(1):125--127.

\bibitem[Arlot and Celisse, 2010]{arlot2010survey}
Arlot, S. and Celisse, A. (2010).
\newblock A survey of cross-validation procedures for model selection.
\newblock {\em Statistics Surveys}, 4:40--79.

\bibitem[Arlot and Celisse, 2011]{arlot2011segmentation}
Arlot, S. and Celisse, A. (2011).
\newblock Segmentation of the mean of heteroscedastic data via
  cross-validation.
\newblock {\em Statistics and Computing}, 21(4):613--632.

\bibitem[Arlot and Lerasle, 2016]{arlot2016choice}
Arlot, S. and Lerasle, M. (2016).
\newblock Choice of {$V$} for {$V$}-fold cross-validation in least-squares
  density estimation.
\newblock {\em The Journal of Machine Learning Research}, 17(1):7256--7305.

\bibitem[Baraud, 2011]{baraud2011estimator}
Baraud, Y. (2011).
\newblock Estimator selection with respect to hellinger-type risks.
\newblock {\em Probability theory and related fields}, 151(1-2):353--401.

\bibitem[Baraud et~al., 2014]{baraud2014estimator}
Baraud, Y., Giraud, C., and Huet, S. (2014).
\newblock Estimator selection in the {G}aussian setting.
\newblock 50(3):1092--1119.

\bibitem[Breiman et~al., 1984]{breiman2017classification}
Breiman, L., Friedman, J.~H., Olshen, R.~A., and Stone, C.~J. (1984).
\newblock {\em Classification and regression trees}.
\newblock Wadsworth Statistics/Probability Series. Wadsworth Advanced Books and
  Software, Belmont, CA.

\bibitem[Burman, 1989]{burman1989comparative}
Burman, P. (1989).
\newblock A comparative study of ordinary cross-validation, $v$-fold
  cross-validation and the repeated learning-testing methods.
\newblock {\em Biometrika}, 76(3):503--514.

\bibitem[Burman, 1990]{burman1990estimation}
Burman, P. (1990).
\newblock Estimation of optimal transformations using $v$-fold cross validation
  and repeated learning-testing methods.
\newblock {\em Sankhy{\=a}: The Indian Journal of Statistics, Series A},
  52(3):314--345.

\bibitem[Celisse, 2014]{celisse2014optimal}
Celisse, A. (2014).
\newblock Optimal cross-validation in density estimation with the ${L}^2$-loss.
\newblock {\em The {A}nnals of Statistics}, 42(5):1879--1910.

\bibitem[Craven and Wahba, 1978]{craven1978smoothing}
Craven, P. and Wahba, G. (1978).
\newblock Smoothing noisy data with spline functions.
\newblock {\em Numerische mathematik}, 31(4):377--403.

\bibitem[Ding et~al., 2018]{ding2018model}
Ding, J., Tarokh, V., and Yang, Y. (2018).
\newblock Model selection techniques: An overview.
\newblock {\em IEEE Signal Processing Magazine}, 35(6):16--34.

\bibitem[Fan and Peng, 2004]{fan2004nonconcave}
Fan, J. and Peng, H. (2004).
\newblock Nonconcave penalized likelihood with a diverging number of
  parameters.
\newblock {\em The {A}nnals of Statistics}, 32(3):928--961.

\bibitem[Feng and Yu, 2019]{Feng2019}
Feng, Y. and Yu, Y. (2019).
\newblock The restricted consistency property of leave-$n_v$-out
  cross-validation for high-dimensional variable selection.
\newblock {\em Statistica Sinica}, 29(3):1607--1630.

\bibitem[Geisser, 1975]{geisser1975predictive}
Geisser, S. (1975).
\newblock The predictive sample reuse method with applications.
\newblock {\em Journal of the American Statistical Association},
  70(350):320--328.

\bibitem[Harrison and Rubinfeld, 1978]{harrison1978hedonic}
Harrison, D. and Rubinfeld, D.~L. (1978).
\newblock Hedonic housing prices and the demand for clean air.
\newblock {\em Journal of Environmental Economics and Management},
  5(1):81--102.

\bibitem[Lei, 2020]{Lei2019}
Lei, J. (2020).
\newblock Cross-validation with confidence.
\newblock {\em Journal of the American Statistical Association},
  115(532):1978--1997.

\bibitem[Li, 1984]{li1984consistency}
Li, K.-C. (1984).
\newblock Consistency for cross-validated nearest neighbor estimates in
  nonparametric regression.
\newblock {\em The Annals of Statistics}, 12(1):230--240.

\bibitem[Li, 1987]{li1987asymptotic}
Li, K.-C. (1987).
\newblock Asymptotic optimality for $ {C}_p, {C}_l $, cross-validation and
  generalized cross-validation: discrete index set.
\newblock {\em The Annals of Statistics}, 15(3):958--975.

\bibitem[Maillard et~al., 2021]{maillard2017cross}
Maillard, G., Arlot, S., and Lerasle, M. (2021).
\newblock Aggregated hold-out.
\newblock {\em Journal of Machine Learning Research}, 22(20):1--55.

\bibitem[Nadaraya, 1964]{nadaraya1964estimating}
Nadaraya, E.~A. (1964).
\newblock On estimating regression.
\newblock {\em Theory of Probability \& Its Applications}, 9(1):141--142.

\bibitem[Nemirovski, 2000]{nemirovski2000topics}
Nemirovski, A. (2000).
\newblock Topics in non-parametric statistics.
\newblock {\em {L}ectures on {P}robability {T}heory and {S}tatistics
  ({S}aint-{F}lour 1998)}, 1738:85--277.

\bibitem[Picard and Cook, 1984]{picard1984cross}
Picard, R.~R. and Cook, R.~D. (1984).
\newblock Cross-validation of regression models.
\newblock {\em Journal of the American Statistical Association},
  79(387):575--583.

\bibitem[Racine, 2000]{racine2000consistent}
Racine, J. (2000).
\newblock Consistent cross-validatory model-selection for dependent data:
  $hv$-block cross-validation.
\newblock {\em Journal of Econometrics}, 99(1):39--61.

\bibitem[Shao, 1993]{shao1993linear}
Shao, J. (1993).
\newblock Linear model selection by cross-validation.
\newblock {\em Journal of the American Statistical Association},
  88(422):486--494.

\bibitem[Shao, 1997]{shao1997asymptotic}
Shao, J. (1997).
\newblock An asymptotic theory for linear model selection (with discussion).
\newblock {\em Statistica Sinica}, 7(2):221--242.

\bibitem[Speckman, 1985]{speckman1985spline}
Speckman, P. (1985).
\newblock Spline smoothing and optimal rates of convergence in nonparametric
  regression models.
\newblock {\em The Annals of Statistics}, 13(3):970--983.

\bibitem[Stone, 1982]{stone1982optimal}
Stone, C.~J. (1982).
\newblock Optimal global rates of convergence for nonparametric regression.
\newblock {\em The {A}nnals of Statistics}, 10(4):1040--1053.

\bibitem[Stone, 1974]{stone1974cross}
Stone, M. (1974).
\newblock Cross-validatory choice and assessment of statistical predictions
  (with discussion).
\newblock {\em Journal of the Royal Statistical Society. Series B},
  36(2):111--147.

\bibitem[Wong, 1983]{wong1983consistency}
Wong, W.~H. (1983).
\newblock On the consistency of cross-validation in kernel nonparametric
  regression.
\newblock {\em The Annals of Statistics}, 11(4):1136--1141.

\bibitem[Yang, 2006]{yang2006comparing}
Yang, Y. (2006).
\newblock Comparing learning methods for classification.
\newblock {\em Statistica Sinica}, 16(2):635--657.

\bibitem[Yang, 2007]{yang2007consistency}
Yang, Y. (2007).
\newblock Consistency of cross validation for comparing regression procedures.
\newblock {\em The Annals of Statistics}, 35(6):2450--2473.

\bibitem[Yang, 2008]{yang2008localized}
Yang, Y. (2008).
\newblock Localized model selection for regression.
\newblock {\em Econometric Theory}, 24(2):472--492.

\bibitem[Zhan and Yang, 2022]{zhan2021profile}
Zhan, Z. and Yang, Y. (2022).
\newblock Profile electoral college cross-validation.
\newblock {\em Information Sciences}, 586:24--40.

\bibitem[Zhang, 2010]{zhang2010nearly}
Zhang, C.-H. (2010).
\newblock Nearly unbiased variable selection under minimax concave penalty.
\newblock {\em The {A}nnals of Statistics}, 38(2):894--942.

\bibitem[Zhang, 1993]{zhang1993model}
Zhang, P. (1993).
\newblock Model selection via multifold cross validation.
\newblock {\em The Annals of Statistics}, 21(1):299--313.

\bibitem[Zhang and Yang, 2015]{zhang2015cross}
Zhang, Y. and Yang, Y. (2015).
\newblock Cross-validation for selecting a model selection procedure.
\newblock {\em Journal of Econometrics}, 187(1):95--112.

\end{thebibliography}
\end{document}